%% file: main.tex
\documentclass{article}
\usepackage{arxiv}
\input{packages}
\input{macros}

\title{Diffusion Models for Inverse Problems}

\author{
  Hyungjin Chung \\
  EverEx \\
  \texttt{hj.chung@everex.co.kr} \\
  \And
  Jeongsol Kim \\
  KAIST \\
  \texttt{jeongsol@kaist.ac.kr} \\
  \And
  Jong Chul Ye \\
  KAIST \\
  \texttt{jong.ye@kaist.ac.kr} \\
}

\begin{document}

\maketitle

\begin{abstract}
    Using diffusion priors to solve inverse problems in imaging have significantly matured over the years. In this chapter, we review the various different approaches that were proposed over the years. We categorize the approaches into the more classic explicit approximation approaches and others, which include variational inference, sequential monte carlo, and decoupled data consistency. We cover the extension to more challenging situations, including blind cases, high-dimensional data, and problems under data scarcity and distribution mismatch. More recent approaches that aim to leverage multimodal information through texts are covered. Through this chapter, we aim to (i) distill the common mathematical threads that connect these algorithms, (ii) systematically contrast their assumptions and performance trade‑offs across representative inverse problems, and (iii) spotlight the open theoretical and practical challenges by clarifying the landscape of diffusion model based inverse problem solvers.
\end{abstract}

\section{Introduction}

We consider inverse problems in the following form
\begin{align}
\label{eq:ip_base}
    \y = \Ac(\x) + \n
\end{align}
where $\Ac: \Rd^n \mapsto \Rd^m, m < n$ is the forward operator that maps the signal that we wish to recover, $\x \in \Rd^n$, to the measurement $\y \in \Rd^m$, and the process is corrupted by noise $\n \in \Rd^m$\footnote{A special yet widely used case is where we have a linear measurement. In such cases, we express $\Ac = A \in \Rd^{m \times n}$.}.
Due to the ill-posedness of the problem, infinitely many feasible solutions exist, and perfect recovery is impossible~\citep{tarantola2005inverse}. Among the feasible solutions, we aim to find a {\em good} set of solutions that also match the characteristics of the real-world data. Mathematically, this can be handily written down with Bayes rule
\begin{align}
\label{eq:bayes}
    p(\x|\y) = p(\x)p(\y|\x)/p(\y), \quad p(\y) = \int_\x p(\y|\x).
\end{align}
One of the most widely studied and used cases is when the likelihood function is a Gaussian model, i.e. $p(\y|\x) = \Nc(\y; \Ac(\x), \sigma_\y^2 I)$. It is easy to see that this corresponds to the case where $\n = \sigma_y\epsilonb,\,\epsilonb \sim \Nc(0, I)$.

Due to the nature of the problem, it is up to the user to define the type of recovery one wants. The following three are among the most widely opted goals:
\begin{enumerate}
    \item Sampling from the posterior (i.e. posterior sampling): $\x \sim p(\x|\y)$
    \item Finding a minimum mean-squared error (MMSE) estimate: $\x = \Ed[\x|\y]$
    \item Finding a maximum a posteriori (MAP) estimate: $\x = \argmax_\x p(\x|\y)$
\end{enumerate}
\cite{blau2018perception} shows that there is a trade-off between perception and distortion, and one cannot maximize perception and minimize distortion at the same time\footnote{Nevertheless, \cite{jalal2021robust} shows that posterior sampling is nearly optimal in terms of distortion.}. 
Note that any of the above goals can be solved by specifying the posterior, which, in turn, can be naturally achieved by specifying the prior. All inverse problem solvers, either explicitly or implicitly, uses this prior function. In this work, we focus mostly on posterior sampling methods that leverage the {\em generative} prior~\citep{bora2017compressed}, in the sense that the prior function is defined through a deep generative model that is trained from data sources.

In the modern generative AI era, modeling the prior data distribution through a generative model is becoming ever more powerful and prominent. Among them, diffusion models~\citep{ho2020denoising,song2020score} have become the predominant paradigm in modeling the distribution of images and videos. While there are more recent variants of diffusion models such as flow matching~\citep{lipman2023flow}, rectified flow~\citep{liu2023flow}, etc., we simply refer to them as diffusion models hereafter as the principles remain the same\footnote{They are indeed equivalent in the case where the reference distribution is a Gaussian~\citep{gao2025diffusionmeetsflow}.}.

As directly modeling the distribution is hard due to the existence of the normalization constant, a clever bypass is to learn the gradient of the log density $\nabla_\x \log p(\x)$, often called the score function~\citep{hyvarinen2005estimation}. Diffusion models learn a family of blurred score functions $\nabla_{\x_t} \log p(\x_t)$ in various noise levels $t \in [0, T]$, with $t = 0$ corresponding to the original data distribution, and $t = T$ resulting in the reference Gaussian distribution. Once the diffusion model is trained along this forward diffusion trajectory, one can sample from the learned distribution by running a reverse diffusion trajectory, which can be characterized by a stochastic differential equation (SDE), or equivalently, an ordinary differential equation (ODE), in the continuous time limit~\citep{song2020score}.

As the reverse diffusion process involves the score function of the prior, we are able to sample from the posterior if we use the score function of the posterior
\begin{align}
\label{eq:nabla_bayes}
    \nabla_{\x_t} \log p(\x_t|\y) = \nabla_{\x_t} \log p(\x_t) + \nabla_{\x_t} \log p(\y|\x_t).
\end{align}
While this may sound straightforward, $p(\y|\x_t)$ is in fact, intractable, and hence requires some form of approximation, or other ways to bypass the computation. In this chapter, we review some of the most widely used Diffusion model based Inverse problem Solvers (DIS) by comparing the categorizing the methods into the ones that make explicit approximations to this term, and other approaches. We note that \cite{daras2024survey} provides a comprehensive review and taxonomy of existing DIS, and we reuse parts of their layout for ease of comparison. However, our chapter diverges by identifying new classes, pushing the timeline to mid-2025, and covering other extensions (e.g. high-dimensional data).

This chapter is structured as follows: In Sec.~\ref{sec:background_diffusion}, we review the fundamentals of diffusion models in both the score-perspective and the variational perspective. In Sec.~\ref{sec:explicit}, we study the explicit approximation methods, with a focus on diffusion posterior sampling~\citep{chung2023diffusion}. In Sec.~\ref{sec:others}, we review a taxonomy of DIS that does not belong to the explicit category, but offers other principled approaches. In Sec.~\ref{sec:extension}, we extend the solvers to more challenging situations, e.g. blind inverse problems. In Sec.~\ref{sec:text}, we review approaches that leverage texts as additional source of control knob to deduce solutions. Finally, in Sec.~\ref{sec:discussion_and_conclusion}, we conclude by discussing the current status and future perspectives of DIS.

\section{Background: Diffusion Models}
\label{sec:background_diffusion}

\subsection{Score perspective}
\label{subsec:score_perspective}

Consider the continuous diffusion process $\x_t, t\in[0,T]$ with $\x_t\in \Rd^d$~\citep{song2020score}. We initialize the process with $\x_0 \sim p_0(\x)$, where $p_0 = p_{\text{data}}$ represents our initial data distribution, and let $\x_T \sim p_T$, with $p_T$ being a reference distribution from which we can draw samples. The forward noising process spanning from $t = 0 \rightarrow T$ is characterized by the following It$\hat{\text{o}}$ stochastic differential equation: 
\begin{equation} 
\label{eq:forward-sde} 
    d\x_t = \f(\x_t, t)dt + g(t)d\w, \quad \f: \Rd^d \times \Rd \mapsto \Rd^d,\, g: \Rd \mapsto \Rd, 
\end{equation} 
where $\f$ denotes the drift function associated with $\x_t$, and $g$ signifies the diffusion coefficient linked with the standard $d$-dimensional Brownian motion $\w \in \Rd^d$. Through the judicious selection of $\f$ and $g$, one can asymptotically converge towards the Gaussian distribution as $t \to T$. When the drift function $\f$ is defined as an affine function of $\x$, specifically $\f(\x, t) = f(t)\x$, it follows that the perturbation kernel $p(\x_t|\x_0)$ consistently exhibits Gaussian characteristics, with its parameters being derivable in closed-form. Consequently, the process of perturbing the data utilizing the perturbation kernel $p(\x_t|\x_0)$ can be accomplished without the necessity of executing the forward SDE.

For the specified forward SDE in \eqref{eq:forward-sde}, it can be demonstrated that a corresponding reverse-time SDE exists, which operates in a backward manner~\citep{song2020score,huang2021variational,anderson1982reverse}: 
\begin{align} 
\label{eq:reverse-sde} 
    d\x_t &= [\f(\x_t, t) - g(t)^2 \nabla_{\x_t} \log p_t(\x_t)]dt + g(t) d\bar{\w} 
\end{align} 
where $dt$ represents the infinitesimal {\em negative} time increment, and $\bar{\w}$ is the standard Brownian motion progressing in reverse. Executing the reverse diffusion as delineated in \eqref{eq:reverse-sde} by initializing with a random Gaussian noise would facilitate sampling from $p_0(\x)$. It is evident that access to the time-conditional score function $\nabla_{\x_t} \log p_t(\x_t)$ is requisite, which corresponds to the score function of the smoothed data distribution that has been convolved with a Gaussian kernel.

An intriguing observation is that there exists a corresponding deterministic ordinary differential equation (ODE) associated with \eqref{eq:reverse-sde}, which is expressed as 
\begin{align} 
\label{eq:pf-ode} 
    d\x_t &= [\underbrace{\f(\x_t, t) - \frac{1}{2}g(t)^2 \nabla_{\x_t} \log p_t(\x_t)}_{=: \tilde\f_\theta(\x_t, t)}]dt. 
\end{align} 
The ODE represented in \eqref{eq:pf-ode} is referred to as the probability-flow ODE (PF-ODE). While both \eqref{eq:reverse-sde} and \eqref{eq:pf-ode} yield the same law $p_t(\x_t)$, PF-ODE possesses several notable properties. Firstly, diffusion models may be reconceptualized as a variant of continuous normalizing flows (CNF)~\citep{chen2018neural} by interpreting the network as $\tilde\f_\theta$, thereby facilitating tractable likelihood computations. Secondly, ODE solvers generally exhibit superior behavior in comparison to SDE solvers. Utilizing the PF-ODE instead of the reverse SDE results in expedited sampling.

It is feasible to train a neural network to approximate the true score function through score matching~\citep{hyvarinen2005estimation}, thereby estimating $\s_{\theta}(\x_t, t) \approx \nabla_{\x_t} \log p_t(\x_t)$, which can subsequently be incorporated into \eqref{eq:reverse-sde}. Nonetheless, it is acknowledged that the application of either explicit or implicit score matching poses significant challenges in terms of scalability, primarily due to inherent instabilities and substantial computational demands. To address these technical obstacles, denoising score matching (DSM) is employed: 
\begin{align} 
\label{eq:dsm} 
    \theta^* = \argmin_\theta \Ed_{t \sim {\rm Unif}(0, T), \x_t \sim p(\x_t|\x_0), \x_0 \sim p(\x_0)}\left[\|\s_\theta(\x_t, t) - \nabla_{\x_t} \log p(\x_t|\x_0)\|_2^2\right]. 
\end{align} 
It is pertinent to acknowledge that DSM is fundamentally equivalent to the training of a denoising autoencoder (DAE) across various noise levels~\citep{vincent2011connection}, which are dictated by an auxiliary input $t$. Specifically, let us examine the most basic forward perturbation kernel defined as $p(\x_t|\x_0) = \Nc(\x_t; \x_0, t^2\Ib)$. By establishing a denoiser parametrization $D_\theta(\x_t, t) \triangleq -\s_\theta(\x_t, t) / t^2$, it becomes evident that \eqref{eq:dsm} can be reformulated as: 
\begin{align} 
\label{eq:dae} 
    \theta^* = \argmin_\theta \Ed_{t \sim {\rm Unif}(0, T), \x_t \sim p(\x_t|\x_0), \x_0 \sim p(\x_0)}\left[t\|D_\theta(\x_t, t) - \x_0\|_2^2\right]. 
\end{align} 
The correspondence between \eqref{eq:dsm} and \eqref{eq:dae} is also fundamentally linked to Tweedie's theorem~\citep{efron2011tweedie}.

\begin{restatable}[Tweedie's theorem]{theorem}{tweedie} 
\label{thm:tweedie} 
    In the context of a Gaussian perturbation kernel represented as $p(\x_t|\x_0) = \Nc(\x_t; s_t\x_0, \sigma_t^2\Ib)$, the posterior mean is articulated mathematically as: \begin{align} \Ed[\x_0|\x_t] = \frac{1}{s_t}(\x_t + \sigma_t^2 \nabla_{\x_t} \log p(\x_t)) 
\end{align} 
\end{restatable}

In essence, the parametrization delineated in \eqref{eq:dae} serves as a direct means of estimating the posterior mean $\Ed[\x_0|\x_t]$. Irrespective of the chosen parametrization, and due to the implications of Theorem~\ref{thm:tweedie}, diffusion models can be conceptualized as possessing two complementary representations: the noisy variable $\x_t$, which evolves according to the reverse SDE outlined in \eqref{eq:reverse-sde}, and the posterior mean $\Ed[\x_0|\x_t]$, which is implicitly characterized by Tweedie's theorem and may be interpreted as the terminal point of the trajectory when adopting a tangent direction relative to the current step.

By choosing $s_t = 1, \sigma_t = t$, the PF-ODE reads
\begin{align}
\label{eq:pf_ode_simple}
    d\x_t = -t\nabla_{\x_t}\log p(\x_t) = \frac{\x_t - \Ed[\x_0|\x_t]}{t}\, dt
\end{align}

\subsection{Variational perspective}
\label{subsec:variational_perspective}

Parallel to the evolution of the score-based framework concerning diffusion models, a variational framework was concurrently established~\citep{sohl2015deep,ho2020denoising}, which now forges a connection between diffusion models and Variational Autoencoders (VAEs)~\citep{kingma2013auto}. More specifically, within this framework, diffusion models are conceptualized as a hierarchical latent variable model referred to as denoising diffusion probabilistic models (DDPM) 
\begin{align} 
\label{eq:ddpm_generative} 
    p_\theta(\x_0) = \int p_\theta(\x_T)\prod_{t=1}^T p_\theta^{(t)}(\x_{t-1}|\x_t)\,d\x_{1:T}, 
\end{align} 
where $\x_{\{1,\dots, T\}} \in \Rd^d$. The neural network that characterizes $p_\theta$ is subsequently optimized by minimizing the evidence lower bound (ELBO) 
\begin{align} 
\label{eq:elbo} 
    \Ed[-\log p_\theta(\x_0)] \leq \Ed_q\left[ - \log \frac{p_\theta(\x_{0:T})}{q(\x_{1:T}|\x_0)} \right] = \Ed_q\left[ -\log p(\x_T) - \sum_{t\geq 1}\log \frac{p_\theta(\x_{t-1}|\x_t)}{q(\x_t|\x_{t-1})} \right] 
\end{align} where the inference distribution $q$ is delineated by the Markovian forward conditional densities 
\begin{align} 
\label{eq:ddpm_forward_single} 
    q(\x_t|\x_{t-1}) &= \Nc(\x_t|\sqrt{\beta_t} \x_{t-1}, (1 - \beta_t) I), \\ 
    q(\x_t|\x_{0}) &= \Nc(\x_t|\sqrt{\bar\alpha_t} \x_0, (1 - \bar\alpha_t) I). \label{eq:ddpm_forward} 
\end{align} 
In this context, the noise schedule $\beta_t$ is characterized as an increasing sequence indexed by $t$, with $\bar\alpha_t := \prod_{i=1}^t \alpha_t,\, \alpha_t := 1 - \beta_t$. The selection of the noise schedule is made such that the signal coefficient $\sqrt{\bar\alpha_t}$ approaches 0 as $t \rightarrow T$, thereby ensuring that the noise coefficient $1 - \bar\alpha_t$ approaches 1, thereby converging towards the standard normal distribution. In contrast to the VE diffusion choice elaborated in Sec.~\ref{subsec:score_perspective}, the selection employed here is denoted as variance preserving (VP). Notably, the discrete VP configuration in \eqref{eq:ddpm_forward_single}, when transitioned to its continuous analogue by increasing the number of discretization steps to $N \rightarrow \infty$, engenders the following Stochastic Differential Equation (SDE) 
\begin{align} 
\label{eq:vpsde} 
    d\x = -\frac{1}{2}\beta_t\x\,dt + \sqrt{\beta_t}d\w. 
\end{align} 
The minimization of the ELBO objective in \eqref{eq:elbo} fundamentally gives rise to the following optimization challenge 
\begin{align} 
\label{eq:variational_kl_minimization} 
    \min_\theta \Ed_q\left[\sum_{t>1} D_{\rm KL}(q(\x_{t-1}|\x_t,\x_0) || p_\theta(\x_{t-1}|\x_t))\right]. 
\end{align} 
The KL minimization task delineated in \eqref{eq:variational_kl_minimization} is computationally feasible as both distributions are Gaussian. For the initial term, this derives from the application of Bayes' rule alongside the Markov property 
\begin{align} 
    q(\x_{t-1}|\x_t, \x_0) &= q(\x_t|\x_{t-1}, \x_0) \frac{q(\x_{t-1}|\x_0)}{q(\x_t|\x_0)} = \Nc(\x_{t-1}; \tilde\mub_t(\x_t,\x_0), \tilde\beta_t\Ib), \\ 
    \mbox{where} \quad \tilde\mub_t(\x_t,\x_0) &:= \frac{\sqrt{\bar\alpha_{t-1}}\beta_t}{1 - \bar\alpha_t}\x_0 + \frac{\sqrt{\alpha_t} (1 - \bar\alpha_{t-1})}{1 - \bar\alpha_t}\x_t,\, \tilde\beta_t := \frac{1 - \bar\alpha_{t-1}}{1 - \bar\alpha_t} \beta_t. 
\end{align} 
For the subsequent term, the reverse distribution is Gaussian as we account for minimal perturbations pertinent to a singular step of forward diffusion~\citep{ho2020denoising}. A common parametrization is established as follows 
\begin{align} 
\label{eq:ddpm_param} 
    &p_\theta(\x_{t-1}|\x_t) = \Nc(\x_{t-1}; \mub_\theta(\x_t, t), \tilde\beta\Ib), \\ 
    &\mbox{where} \quad \mub_\theta(\x_t,t) = \frac{1}{\sqrt{\alpha_t}}\left( \x_t - \frac{\beta_t}{\sqrt{1 - \bar\alpha_t}}\epsilonb_\theta(\x_t,t) \right). 
\end{align} 
Under this formulation, the ELBO objective in \eqref{eq:elbo} can be streamlined to the epsilon-matching objective by disregarding the time-dependent weighting factors 
\begin{align} 
\label{eq:epsilon_matching} 
    \theta^* = \argmin_\theta \Ed_{\x_t \sim q(\x_t|\x_0), \x_0 \sim p_{\rm data}(\x_0), \epsilonb \sim \Nc(0, \Ib)}\left[\|\epsilonb_\theta(\x_t, t) - \epsilonb\|_2^2\right]. 
\end{align} 
Epsilon matching is fundamentally analogous to the DSM/DAE objective in \eqref{eq:dsm}, \eqref{eq:dae}, differing solely by a constant with an alternative parametrization. Given the correspondence between the forward noising distribution in \eqref{eq:vpsde} and the learning objective in \eqref{eq:dsm},\eqref{eq:epsilon_matching}, it becomes evident that the two frameworks essentially converge upon the same model.

Inference can be executed by incorporating the trained $\epsilonb_\theta$ to approximate the expectation of $p_\theta(\x_{t-1}|\x_t)$, culminating in the subsequent iterative expression 
\begin{align} 
\label{eq:ddpm_inference} 
    \x_{t-1} = \frac{1}{\sqrt{\alpha_t}}\left( \x_t - \frac{\beta_t}{\sqrt{1 - \bar\alpha_t}}\epsilonb_\theta(\x_t,t) \right) + \tilde\beta_t\epsilonb, \quad \epsilonb \sim \Nc(\bm{0}, \Ib). 
\end{align} 
It is noteworthy that analogous to the reverse stochastic differential equation (SDE) delineated in \eqref{eq:reverse-sde}, stochastic perturbations are incorporated in each iteration throughout the DDPM sampling process, resulting in a protracted inference duration. A conventional methodology to mitigate this phenomenon, akin to the transition towards the PF-ODE, is facilitated by denoising diffusion implicit models (DDIM)~\citep{song2020score}, wherein an alternative inference distribution is proposed 
\begin{align} 
\label{eq:ddim} 
    q_\eta(\x_{t-1}|\x_t,\x_0) = \Nc(\x_{t-1}; \sqrt{\bar\alpha_{t-1}}\x_0 + \sqrt{1 - \bar\alpha_{t-1} - \eta\tilde\beta_t^2}\frac{\x_t - \sqrt{\bar\alpha_t}\x_0}{\sqrt{1 - \bar\alpha_t}}, \eta\tilde\beta_t^2\Ib), 
\end{align} 
where $\eta \in [0, 1]$. By establishing $\eta = 1.0$, the original DDPM sampling procedure is reinstated with maximal stochasticity. Conversely, by designating $\eta = 0.0$, a deterministic sampling mechanism is achieved, which can be demonstrated to be equivalent to the variance preserving PF-ODE~\citep{song2020score}. Employing diminished values of $\eta$ tends to yield superior outcomes when the objective is to minimize the number of function evaluations (NFE).

%%%%%%%%%%%%%%%%%%%%%%%%% 25.06.29 %%%%%%%%
% 3단원 / 우리 main - 형진
% 1. pointwise estimation
% explicit estimation

% 4단원 / others
% 1. variational inference - 정솔
% DAVI / APS
% training / -free (amortized?)
% map red-diff / flair
% 2. DAPS 계열 - 형진
% 3. SMC - 형진

% 5단원 / blind, 3d, ood 뭉뚱그려서 application들 얘기하자 - 형진

% 6단원 / text - 정솔

% 7단원 / fast solver (CM, bridge) - 정솔
%%%%%%%%%%%%%%%%%%%%%%%%%%%%%%%%%%%%%%%%%%%%%%

\section{Explicit approximation methods}
\label{sec:explicit}

Many of the earlier works that aimed to solve inverse problems with diffusion models, whether explicitly mentioned in the original work or not, can be perceived as explicit approximation methods for the time-dependent log-likelihood $p(\y|\x_t)$ in \eqref{eq:nabla_bayes}. In this section, we review some of the canonical works that belong to this category, with a specific focus on the DPS~\citep{chung2023diffusion} family. 

The first works that used diffusion model-like annealing-denoising steps with projection-like data consistency steps were \cite{song2019generative,kadkhodaie2021stochastic}. While the details differ, one can understand the algorithms as alternating the denoising step and the data consistency projection step, gradually decreasing the noise level, starting from pure Gaussian noise. Note that the earlier works mostly focused on linear inverse problems, where $\Ac = A$.

\paragraph{Score-ALD~\citep{jalal2021robust}}
In this work, the authors focused on the task of compressed-sensing MRI, where the following approximation was used
\begin{align}
\label{eq:score_ald}
    \nabla_{\x_t} \log p(\y|\x_t) \approx -\frac{A^\top (\y - A\x_t)}{\sigma_y^2 + \gamma_t^2},
\end{align}
where $\gamma_t$ was set to be a hyperparameter that decays as $t$ approaches 0. 

\paragraph{Score-SDE~\citep{song2020score}}
Score-SDE focused on linear inverse problems with an orthogonal matrix $A$
\begin{align}
\label{eq:score_sde}
    \nabla_{\x_t} \log p(\y|\x_t) \approx -A^\top (\y + \sigma_t\epsilonb - A\x_t),
\end{align}
which corresponds to a noisy projection onto $\y + \sigma_t\epsilonb = A\x_t$.

\subsection{DDRM family}
\label{subsec:DDRM family}

The methods that belong to this category explicitly uses singular value decomposition (SVD) $A = U\Sigma V^\top, U \in \Rd^{m \times m}, V \in \Rd^{n \times n}, \Sigma \in \Rd^{m \times n}$, with $\Sigma$ being a rectangular diagonal matrix with singular values $\{s_j\}_{j=1}^m$ as the diagonal elements. Notice that one can then rewrite the linear inverse problem as
\begin{align}
\label{eq:ddrm_family_ip}
    \bar\y = \Sigma \bar\x + \sigma_y \bar\epsilonb, \quad \mbox{where} \quad \bar\y := U^\top \y, \bar\x := V^\top \x, \bar\epsilonb := U^\top \epsilonb.
\end{align}
Once $\bar\x$ is recovered from \eqref{eq:ddrm_family_ip}, $\hat\x = V\bar\x$.

\paragraph{SNIPS~\citep{kawar2021snips}}
The approximation reads
\begin{align}
\label{eq:snips}
    \nabla_{\bar\x_t} \log p(\bar\y|\bar\x_t) \approx -\Sigma^\top \left|\sigma_y^2I - \sigma_t^2\Sigma\Sigma^\top\right|^\dagger \left(\bar\y - \Sigma\bar\x_t\right),
\end{align}
where the gradient points to a direction weighted by the magnitude of the difference between the diffusion noise level $\sigma_t^2$ and the measurement noise $\sigma_y^2$, additionally weighted by the singular values $s_i^2$.

\paragraph{DDRM~\citep{kawar2022denoising}}
DDRM is an extension of SNIPS which incorporates DDIM sampling, an additional {\em mixing} hyperparameter $\eta$, and using the posterior mean $\bar\x_{0|t} := V\Ed[\x_0|\x_t]$
\begin{align}
\label{eq:ddrm}
    \nabla_{\bar\x_t} \log p(\bar\y|\bar\x_t) \approx -\Sigma^\top \left|\sigma_y^2I - \sigma_t^2\Sigma\Sigma^\top\right|^\dagger \left(\bar\y - \Sigma\bar\x_{0|t}\right).
\end{align}
Notice that an element-wise expression of \eqref{eq:ddrm} can be written as
\begin{align}
\label{eq:ddrm_simple}
    p(\bar\vx_t^{(i)}|\vx_{t+1},\vy) = 
    \begin{cases}
        \gN(\bar\vx_t^{(i)}; \bar\vx_{0|t+1}^{(i)}, \sigma_t^2) & \mbox{if  } s_i = 0\\
        \gN(\bar\vx_t^{(i)}; \bar\vx_{0|t+1}^{(i)}, \sigma_t^2) & \mbox{if  } \sigma_t < \frac{\sigma_{\vy}}{s_i}\\
        \gN(\bar\vx_t^{(i)}; \bar\vy^{(i)}, \sigma_t^2 - \frac{\sigma_{\vy}^2}{s_i^2}) & \mbox{if  } \sigma_t \geq \frac{\sigma_{\vy}}{s_i} \\
    \end{cases}.
\end{align}
Analagous to the role of mixing coefficient $\eta$ in DDIM sampling, DDRM introduces a hyper-parameter $\eta \in (0, 1]$ to get
\begin{align}
    p(\bar\vx_t^{(i)}|\vx_{t+1},\vy) = 
    \begin{cases}
        \gN(\bar\vx_t^{(i)}; \bar\vx_{0|t+1}^{(i)} + \sqrt{1 - \eta^2}\sigma_t\frac{\bar\vx_{t+1}^{(i)} - \bar\vx_{0|t+1}^{(i)}}{\sigma_{t+1}}, \eta^2\sigma_t^2) & \mbox{if  } s_i = 0\\
        \gN(\bar\vx_t^{(i)}; \bar\vx_{0|t+1}^{(i)} + \sqrt{1 - \eta^2}\sigma_t\frac{\bar\vy^{(i)} - \bar\vx_{0|t+1}^{(i)}}{\sigma_{\vy}/s_i}, \eta^2\sigma_t^2) & \mbox{if  } \sigma_t < \frac{\sigma_{\vy}}{s_i}\\
        \gN(\bar\vx_t^{(i)}; \bar\vy^{(i)}, \sigma_t^2 - \frac{\sigma_{\vy}^2}{s_i^2}) & \mbox{if  } \sigma_t \geq \frac{\sigma_{\vy}}{s_i} \\
    \end{cases}.
\label{eq:ddrm}
\end{align}

\subsection{DPS family}
\label{subsec:dps}

\paragraph{DPS~\citep{chung2023diffusion}}

Notice that
\begin{align}
\label{eq:time_likelihood_decomp}
    p(\y|\x_t) = \int p(\y|\x_0) p(\x_0|\x_t) d\x_0 = \Ed_{\x_0 \sim p(\x_0|\x_t)}[p(\y|\x_0)].
\end{align}
The computation of $p(\x_0|\x_t)$ is challenging, as we would have to marginalize over all the latent steps $t$ through $0$, not to mention the integration over the trajectories. It would be computationally intractable to compute this value every time we need access to the time-conditional likelihood. The idea of DPS is to push the expectation inside
\begin{align}
\label{eq:dps}
    p(\y|\x_t) \approx p(\y|\hat\x_{0|t}), \quad \mbox{where} \quad \hat\x_{0|t} = \Ed[\x_0|\x_t].
\end{align}
This approximation is often referred to as Jensen’s approximation, whose approximation bound has been shown to be controllable in the context of Gaussian measurement scenarios \citep{chung2023diffusion}.
Recall from Theorem~\ref{thm:tweedie} that one can easily compute the MMSE estimate $\hat\x_{0|t}$ through a single forward pass through the score function. From the definition of the forward model, it is then easy to see that
\begin{align}
\label{eq:dps_nabla}
    \nabla_{\x_t} \log p(\y|\hat\x_{0|t}) = -\frac{1}{2\sigma_y^2}\nabla_{\x_t}\|\y - \Ac(\hat\x_{0|t})\|_2^2,
\end{align}
where in practice, an empirical static step $\rho$ size is often employed in the place of $1/2\sigma_y^2$.
The computation of the gradient can be done through backpropagation, as it involves a backward pass through the score function. It is important to note that DPS is fully general in that it is capable of solving non-linear inverse problems with arbitrary forward models if it can be defined.

\paragraph{$\Pi$GDM~\citep{song2023pseudoinverseguided}}
From \eqref{eq:dps}, DPS can be interpreted as using the following approximation
\begin{align}
    p(\x_0|\x_t) \approx \delta(\x_0 - \hat\x_{0|t}).
\end{align}
$\Pi$GDM instead places an isotropic Gaussian distribution for approximation
\begin{align}
    p(\x_0|\x_t) \approx \Nc(\hat\x_{0|t}, r_t^2 I),
\end{align}
where $r_t$ is a hyperparameter. For the case of linear inverse problems, this leads to
\begin{align}
    p(\y|\x_t) \approx \Nc(A\hat\x_{0|t}, r_t^2 AA^\top + \sigma_y^2 I),
\end{align}
and subsequently
\begin{align}
    \nabla_{\x_t} \log p(\y|\x_t) \approx -\nabla_{\x_t} \hat\x_{0|t} (r_t^2 AA^\top + \sigma_y^2 I)^{-1} A^\top (\y - A\hat\x_{0|t})
\end{align}

\paragraph{Moment Matching~\citep{rozet2024learning}}

In moment matching, the authors explicitly calculate the variance matrix for $p(\x_0|\x_t)$, leading to a better approximation
\begin{align}
    p(\x_0|\x_t) \approx \Nc(\hat\x_{0|t}, {\rm Var}[\x_0|\x_t]), \quad \mbox{where} \quad {\rm Var}[\x_0|\x_t] = \sigma_t^2\nabla_{\x_t} \hat\x_{0|t}.
\end{align}
In turn, this leads to
\begin{align}
\label{eq:mm_nabla}
    \nabla_{\x_t} \log p(\y|\x_t) \approx -\nabla_{\x_t}(A\hat\x_{0|t})^\top(\sigma_y^2 I + A\sigma_t^2\nabla_{\x_t}\hat\x_{0|t} A^\top)^{-1} (\y - A\hat\x_{0|t})
\end{align}
Note that in high-dimensions, explicit computation of $\nabla_{\x_t} \hat\x_{0|t}$ is expensive. Nevertheless, Jacobian-vector products (JVP) can be used for efficient computation for both $\Pi$GDM and moment matching.

\paragraph{\cite{peng2024improving}}

In a related work of \cite{peng2024improving}, the authors show that there exists an optimal posterior diagonal posterior covariance in by analyzing the diffusion model under the DDPM framework. The covariance matrix can be determined through maximum likelihood estimation, without relying on the computation of $\nabla_{\x_t}\hat\x_{0|t}$, and it was further shown that using this optimal covariance enhances the performance on robustness in all cases.

\paragraph{DDS~\citep{chung2024decomposed}}

One of the critical downsides of the other methods within the DPS family is that they are slow to compute, and requires excessive memory, since the computation of $\nabla_{\x_t}\hat\x_{0|t}$ is involved. This may not be suitable for large-scale inverse problems, which the authors of \cite{chung2024decomposed} investigate. The key finding of DDS is that, under certain conditions on the data manifold, one can circumvent the heavy computation.

\begin{restatable}[Manifold Constrained Gradient~\citep{chung2022improving}]{proposition}{mcg} 
\label{prop:mcg} 
    Suppose the clean data manifold $\Mc$, where $\x_0$ resides, is represented as an affine subspace and assumes the uniform distribution on $\Mc$. Then,
    \begin{align}
        \frac{\partial \hat\x_{0|t}}{\partial \x_t} &= \frac{1}{\sqrt{\bar\alpha_t}} \Pc_{\Mc} \\
        \hat\x_{0|t} - \gamma_t \nabla_{\x_t} \ell (\hat\x_{0|t}) &= \Pc_\Mc (\hat\x_{0|t} - \xi_t \nabla_{\hat\x_{0|t}}\ell(\hat\x_{0|t})), 
    \end{align}
    for some $\xi_t > 0$, where $\Pc_\Mc$ denotes the orthogonal projection to $\Mc$.
\end{restatable}
This implies that the manifold constrained gradient (MCG) can be regarded as the projected gradient method on the clean data manifold.
To accelerate the convergence of the algorithms, the authors in \cite{chung2024decomposed} proposed performing multiple manifold-constrained update steps following a single neural network function evaluation (NFE) for manifold projection. This approach can be efficiently implemented using the conjugate gradient (CG) method or other Krylov subspace methods, under the assumption that the clean data manifold lies within a Krylov subspace.

\paragraph{Other approaches that improve DPS}

MPGD~\citep{he2024manifold} proposes to {\em project} the DPS gradient to the manifold by leveraging an autoencoder. DSG~\citep{yang2024guidance} imposes a spherical constraint to control the steps reside on the noisy manifold, as discussed in MCG~\citep{chung2022improving}. DMAP~\citep{xu2025rethinking} argues that DPS behaves closer to an MAP estimate rather than a posterior sampler, and thus proposes to make the algorithm behave closer to an MAP approximation method by imposing multi-step gradients, thereby improving performance. DPPS~\citep{wu2024diffusion} reduces variance by proposing multiple candidates at each step of denoising, and only selecting the ones that maximize the data consistency.

\paragraph{Extension to flow models}
Flow-based model~\citep{lipman2023flow} provide a general framework that includes diffusion models as a special case.
FlowChef~\citep{patel2024steering} introduces a general guidance term, $\nabla_{\hat\vx_{0|t}} \ell(\hat\vx_{0|t})$, into the reverse ODE, accompanied by an analysis of error dynamics, and demonstrates it effectiveness across various conditioned image generation tasks including linear inverse problems.
FlowDPS~\citep{kim2025flowdps} extends posterior sampling theory from diffusion models to general affine flows by decomposing a single Euler step into a linear combination of clean and noise estimates, leveraging a generalized Tweedie's formula.

\section{Other methods}
\label{sec:others}

While we focused on explicit approximation methods in Sec.~\ref{sec:explicit}, there exists multiple different categories. In this section, we provide an introduction to some of the widely-acknowledged among them.

\subsection{Variational inference}
\label{subsec:VI}

Another line of work on solving inverse problems with diffusion models stems from variational inference for the posterior distribution $p(\vx|\vy)$. The main advantage of this approach lies in distributional matching, which offers better diversity compared to the DPS family that approximates the log-likelihood at a single sample point $\x_t$ with its MMSE estimate $\hat\x_{0|t}$.
Let $q_\phi(\vx_0|\vy)$ be a variational distribution with parameters $\phi$. The goal of variational inference is to fit $q$ to the target posterior distribution $p$ by minimizing
\begin{align}
\label{eq:vi}
    \min_\phi D_{KL}\left(q_\phi(\vx_0|\vy)\|p(\vx_0|\vy)\right)=\min_\phi \mathbb{E}_{\vx_0\sim q_\phi}\left[\log q_\phi(\vx_0|\vy) - \log p(\vx_0|\vy)\right].
\end{align}
From the definition of KL divergence, and applying Bayes' rule to $p(\vx_0|\vy)$, the objective function is reformulated as
\begin{align}
\label{eq:vi_decompose}
    \min_{\phi} \underbrace{-\mathbb{E}_{\vx\sim q_\phi} \left[\log p(\vy|\vx_0) \right]}_{\text{data consistency}} + \underbrace{D_{KL}(q_\phi(\vx_0|\vy) \| p(\vx_0))}_{\text{regularizer}} + \underbrace{\log p(\vy)}_{\text{constant}}.
\end{align}

\paragraph{RED-Diff~\citep{mardani2023variational}}
Suppose that $q_\phi(\vx_0|\vy)$ is the isotropic Gaussian distribution $\Nc(\vmu, \sigma^2\rmI)$ where $\phi=\{\vmu, \sigma\}$. 
The objective function of \eqref{eq:vi_decompose} is equivalent to
\begin{align}
    \label{eq:vi_reddiff}
    \mathbb{E}_{\vx_0\sim q_\phi(\vx_0|\vy)}\frac{\|\vy-\Ac(\vx_0)\|^2}{2\sigma_y^2}  + \int_0^T \frac{\beta_t}{2} \mathbb{E}_{\vx_t\sim q_\phi(\vx_t|\vy)} \left[\|\nabla_{\vx_t} \log q(\vx_t|\vy) -\nabla_{\vx_t} \log p(\vx_t) \|^2 \right]dt
\end{align}
where $\vx_t \sim q_\phi(\vx_t|\vy)$ denotes the diffusion trajectory computed by forward diffusion process in \eqref{eq:ddpm_forward}. The first term corresponds to data consistency derived from the definition of the forward model, and the second term denotes cumulative different of score functions along $\x_t$ which is derived from the relationship between KL divergence and score matching provided in Theorem 1 of \citep{song2021maximum}. %or single particle approximation for particle-based variational inference as shown in prolific dreamer
% \begin{align}
%     \label{eq:kl_sm}
%     \frac{\partial D_{KL}(q(\vx|\vy)\|p(\vx))}{\partial t} = -\frac{1}{2} \mathbb{E}_{\vx\sim q_t} g^2(t)\left[ \|\nabla_{\vx} \log q(\vx|\vy) - \nabla_{\vx} \log p(\vx)\|^2\right]
% \end{align}
As $q_\phi(\vx_t|\vy)$ is also Gaussian distribution, the optimization problem turns into a stochastic optimization:
\begin{align}
    \label{eq:reddiff}
    \min_{\mub} \frac{\|\vy-\Ac(\mub)\|^2}{2\sigma_y^2} +\mathbb{E}_{t, \epsilonb} \frac{\beta_t(1-\bar\alpha_t)}{\bar\alpha}  \|\epsilonb_\theta(\vx_t,t)-\epsilonb \|^2 
\end{align}
where the variance of $q(\vx_0|\vy)$ is assumed to be a constant near zero so the optimization variable becomes $\mub$.
To improve efficiency and stability, back-propagation through the score network $\theta$ is omitted. Also, time $t$ is sampled from $T$ to $0$ so the solution is reconstructed from coarse semantics to fine details which enhances perceptual quality.
Notably, the second term reduces to the score-distillation loss. From a MAP perspective, the method implements regularization by denoising~\citep{romano2017little}, where a pre-trained denoiser acts as the prior.
Recently, FLAIR~\citep{erbach2025solving} extended this framework to flow-based models by replacing the second term in \eqref{eq:vi_reddiff} with a velocity difference and introducing a trajectory adjustment mechanism to ensure that the intermediate state $\vx_t$, where the score function is evaluated, lies in a high-likelihood region of the marginal distribution $p(\vx_t)$. Specifically, they obtain $\vx_t$ from $\mu$ using forward diffusion sampling that incorporates both deterministic noise, predicted via Tweeedie's formula~\citep{efron2011tweedie, kim2021noise2score, kim2025flowdps}, and stochastic noise.

\paragraph{\cite{feng2023score}}
A uni-modal Gaussian variational family cannot capture a complex, multi-modal posterior distribution. \cite{feng2023score} employ a normalizing flow - RealNVP~\citep{dinh2017density} to represent the variational distribution $q_\phi$. The corresponding objective starts from the same problem,
\begin{align}
\label{eq:berthy}
    \min_\phi D_{KL}(q_\phi(\vx)\|p(\vx|\vy))&=\min_\phi \mathbb{E}_{\vx\sim q_\phi}[\log q_\phi(\vx) - \log p(\vx|\vy)]\\
    &=\min_\phi \mathbb{E}_{\vx\sim q_\phi}[\log q_\phi(\vx)-\log p(\vy|\vx)-\log p(\vx)]
\end{align}
where $-\log p(\vy|\vx)$ is computed analytically from the forward model, $\log p(\vx)$ is approximated with a pre-trained diffusion model $\theta$, and $\log q_\phi(\vx)$ is computationally tractable under RealNVP. 
Unlike methods that merely adjust individual samples toward higher posterior likelihood, this normalizing flow-based formulation allows direct sampling from the learned posterior. Consequently, it avoids hyper-parameter tuning (for example, step sizes for likelihood gradients) and produce diverse, robust samples. The trade-off is higher computational overhead for training and a dependence on the expressive power of the chosen normalizing-flow architecture.

This was later extended to \cite{feng2024variational}, where the computation of $\log p(\x)$ by iterative sampling is replaced with a lower bound that involves the DSM loss, as proposed in \cite{song2021maximum}.

\paragraph{APS~\citep{mammadov2024amortized}}

Notice that \eqref{eq:berthy} requires optimizing $\phi$ for every different observations $\y$, which is costly. Amortized Posterior Sampling (APS) proposes the following amortization
\begin{align}
    \min_\phi D_{KL}(q_\phi(\x_0|\y)\| p_\theta(\x_0|\y)),
\end{align}
which can be implemented as a conditional NF. Specifically, the authors proposed to extend RealNVP to a conditional setting, thereby enabling the use of a single network for all $\y$.

\paragraph{RSLD~\citep{zilberstein2025repulsive}}
% extension to latent diffusion. to remove blur, repulsive regularization.

As another approach to estimate multi-modal posterior distribution, RSLD defines the particle-based variational inference and introduces a repulsive regularization to the score-matching term of \eqref{eq:vi_reddiff}. 
Specifically, it approximate the gradient for minimization problem \eqref{eq:reddiff} with ensemble of gradients:
\begin{align}
\label{eq:rsld_grad}
    \frac{1}{n}\sum_{i=1}^n \nabla_{\mub^{(i)}}\frac{\|\vy-\Ac(\mub^{(i)})\|^2}{2\sigma^2_y}+ \mathbb{E}_{t,\epsilonb} \left[\epsilonb_\theta(\vx_t^{(i)}, t)-\epsilonb - \nabla_{\vx_t^{(i)}}R(\vx_t^{(1)}...\vx_t^{(n)}) \right]
\end{align}
where $\vx_t^{(i)}$ is diffusion trajectory of i-th particle that is computed by forward diffusion process with $\mub^{(i)}$ and $\epsilonb$, $n$ denotes the number of particles, and $R(\vx_t^{(1)},..., \vx_t^{(n)})$ denotes the repulsive regularization defined as
\begin{align}
    \label{eq:repulsive}
    \sum_{j=1}^n \log \left[k(\vx_t^{(i)}, \vx_t^{(j)})\right]^r
\end{align}
This gradient is derived by incorporating ODE of each particles - transformed from variational distribution $q$ to the posterior distribution $p$ via Wasserstein Gradient Flow - into the second term of \eqref{eq:vi_reddiff}.
As a result, RSLD jointly updates $n$ particles using gradient in \eqref{eq:rsld_grad}, yielding diverse samples that follows the posterior distribution.

\paragraph{DAVI~\citep{lee2024diffusion}}
While normalizing flow modesl can represent more complex variational distributions, they typically require multiple iterations to obtain a solution.
DAVI addresses this limitation by training a neural network to estimate $q_\phi(\vx_0|\vy)$ in \eqref{eq:vi_decompose}, enabling one-step sampling.
However, the authors also highlight a challenge, the lack of overlap between the supports of $q_\phi(\vx_0|\vy)$ and $p(\vx_0)$, which leads to unstable training and limited performance.
As a result, DAVI reformulate the problem using integral form of the KL divergence in \eqref{eq:vi_reddiff}. Unlike RedDiff that assumes $q(\vx_t|\vy)$ as Gaussian distribution, DAVI compute $\nabla_{\vx_t} \log q_\phi(\vx_t|\vy)$ by an implicit score function $\vs_\psi$.
Thus, during trainig, DAVI alternates between updaing $q_\phi(\vx_0|\vy)$ and $\vx_\phi$. Specifically, $q_\phi(\vx_0|\vy)$ is updated by minimizing \eqref{eq:vi_reddiff}, using the approximation $\nabla_{\vx_t} \log q_\phi(\vx_t|\vy) \approx \vs_\psi$. In turn, $\vx_\psi$ is trained via denoising score matching using samples from the marginal $q_t(\vx_t|\vy)$, obtained by first drawing $\vx_0 \sim q_\phi(\vx_0|\vy)$ and then applying the forward diffusion process $\vx_t \sim q(\vx_t|\vx_0)$.

\subsection{Decoupled data consistency}
\label{subsec:ddc}

\paragraph{DAPS~\citep{zhang2025improving}}

In explicit approximation methods, the solvers typically alternate between a small step of denoising, and a likelihood gradient step. Often, this results in the resulting samples diverging, especially in challenging cases (e.g. Fourier phase retrieval). One way to mitigate this with more compute, is to leverage more compute. Specifically, rather than relying on the Tweedie estimate as in DPS, one can first run the PF-ODE to sample from $\tilde\x_{0|t}^{(j)} \sim p(\x_0|\x_t)$. Then, to impose data consistency, DAPS runs $N$-step Langevin dynamics
\begin{align}
\label{eq:daps_langevin}
    \tilde\x_{0|t}^{(j+1)} = \tilde\x_{0|t}^{(j)} + \eta_t \left(
    \nabla_{\tilde\x_{0|t}} \log p(\tilde\x_{0|t}^{(j)}|\x_t) +
    \nabla_{\tilde\x_{0|t}} \log p(\y|\tilde\x_{0|t}^{(j)}) + \sqrt{2\eta_t} \epsilonb_j\right),
\end{align}
where $\eta_t > 0$ is a hyperparameter. This process is applied for all $t$, where the next iteration starts with $\x_{t-1} \sim p(\x_{t-1}|\x_0)$.
Such approach {\em decouples} the data consistency with the unconditional sampling steps, i.e. $p(\x_0|\x_t,\y) \propto p(\x_0|\x_t)p(\y|\x_0)$, thereby yielding improved performance in certain challenging cases.

\paragraph{DCDP~\citep{li2024decoupled}}

DCDP follows a similar decoupled approach, but differs in how the data consistency steps are performed. Specifically, \cite{li2024decoupled} proposes to use proximal optimization steps
\begin{align}
\label{eq:dcdp_prox}
    \x_k = \argmin_\x \frac{1}{2}\|\Ac(\x) - \y\|_2^2 + \mu \|\x - \vb_{k-1}\|_2^2,
\end{align}
with $\x$ initialized to $\vb_{k-1}$ at the start of optimization.

\paragraph{SITCOM~\citep{alkhouri2024sitcom}}

SITCOM defines three different criteria in which DIS should satisfy: 1) forward consistency, 2) backward consistency, and 3) measurement consistency. To enable this, akin to CSGM~\citep{bora2017compressed}, optimizes the {\em input} to the diffusion model with
\begin{align}
\label{eq:sitcom}
    \x_t' = \argmin_{\x_t} \|\Ac(D_\theta(\x_t)) - \y\|_2^2,
\end{align}
and additionally imposing proximal constraints as in \eqref{eq:dcdp_prox}. Once the optimization is performed, the sampling steps follow the usual DDIM sampling steps, running \eqref{eq:sitcom} for every $t$ reverse sampling steps. 

\subsection{Sequential Monte Carlo}
\label{subsec:smc}

Sequential Monte Carlo (SMC) methods, also known as particle filters, have emerged as a principled framework for solving inverse problems with diffusion priors. SMC methods enjoys the property that with increased compute (i.e. number of particles $\rightarrow \infty$), the sampler approaches sampling from the {\em true} posterior. The particles, each representing a hypothesis about a solution, are propagated through the reverse diffusion sampling steps, re-weighted according to their consistency with respect to the observation. The algorithms mostly differ on how one constructs the proposal kernel and the reweighting values.

\paragraph{SMCDiff~\citep{trippe2023diffusion}}

SMCDiff aims to construct a scaffold structure given a desired motif, which can be cast as a special case of the noiseless inpainting problem. Specifically, let $\y$ be the motif (i.e. measurement), $\x$ be the scaffold, and $\x = [\y, \z]$, i.e. $\y \in \Rd^m$, $\z \in \Rd^{n-m}$ are sub-vectors of $\x$. Akin to Score-SDE~\citep{song2020score}, one first constructs a forward-diffused motif
\begin{align}
\label{eq:smcdiff_1}
    \y_{1:T} \sim q(\y_{1:T}|\y_0),
\end{align}
which are prepared before the reverse diffusion sampling steps, then cached for later use.
Then, for all the particles that are propagated, the sub-vector that corresponds to the motif are replaced 
\begin{align}
\label{eq:smcdiff_2}
    \forall j, \x_t^{(j)} \gets [\y_{t}, \z_t^{(j)}].
\end{align}
The un-normalized reweighting kernel is then constructed as
\begin{align}
\label{eq:smcdiff_3}
    \forall j, w_t^{(j)} \gets p_\theta(\y_{t-1}|\x_t).
\end{align}
For all reverse diffuison steps and particles, \eqref{eq:smcdiff_1}-\eqref{eq:smcdiff_3} along with the resampling steps are applied.

\paragraph{MCGDiff~\citep{cardoso2024monte}}

MCGDiff first defines $q(\x_t|\y) = \Nc(\x_t; \sqrt{\bar\alpha_t}\y, (1 - \bar\alpha_t))$. The proposal kernel for reverse distribution is defined as
\begin{align}
    p_\theta^\y (\x_t|\x_{t+1}) \propto p_\theta(\x_t|\x_{t+1})q(\x_t|\y).
\end{align}
For every propagated particle, the reweighting kernel is defined as
\begin{align}
    w_t(\x_{t+1}^{(j)}) = \frac{\int p_\theta(\x_t|\x_{t+1}^{(j)})q(\x_t|\y)\,d\x_t}{q(\x_{t+1}^{(j)}|\y)}.
\end{align}
The sampling process follows the usual SMC procedure, with proposal, weighting, and resampling.

\paragraph{FPS~\citep{dou2024diffusion}}

The core technical innovation of FPS is the construction of coupled diffusion process. In addition to the standard sequence of noisy data latents $\x_t$, the algorithm generates a corresponding sequence of noisy measurements $\y_t$, where the noise is correlated to $\x_t$. Specifically, given the forward process
\begin{align}
\label{eq:xt_forward}
    \x_t = a_t\x_{t-1} + b_t\epsilonb_t, \quad \epsilonb_t \sim \Nc(0, I),
\end{align}
one can similarly define
\begin{align}
    \y_0 = \y, \quad \y_t = a_t \y_{t-1} + b_t A\epsilonb_t,
\end{align}
so that $\y_t \sim \Nc(A\x_t, c_t^2\sigma_y^2 I)$, with $c_t = a_1a_2 \dots a_t$. This construction leads to the following closed-form expression
\begin{align}
    p_\theta(\y_{t-1}|\x_{t-1}) = \Nc(A\x_{t-1}, c_{t-1}^2\sigma_y^2 I)
\end{align}
and
\begin{align}
\label{eq:fps_smc_proposal}
    p_\theta(\x_{t-1}|\x_t, \y_{t-1}) \propto p_\theta (\x_{t-1}|\x_t) p_\theta (\y_{t-1}|\x_{t-1}).
\end{align}
FPS uses \eqref{eq:fps_smc_proposal} as the proposal kernel of the SMC procedure, and uses the following resampling weights
\begin{align}
    w_t^{(j)} = \frac{p_\theta(\y_t|\x_t^{(j)}) p_\theta(\x_t^{(j)}|\x_{t-1}^{(j)}) / p_\theta(\x_t^{(j)}|\x_{t+1}^{(j)}, \y_t)}{\sum_{j=1}^M p_\theta(\y_t|\x_t^{(j)}) p_\theta(\x_t^{(j)}|\x_{t+1}^{(j)}) / p_\theta(\x_t^{(j)}|\x_{t+1}^{(j)}, \y_t)}
\end{align}

\paragraph{Connections to inference-time scaling}

\cite{singhal2025general} recently drew connections to inference-time scaling of diffusion models, as SMC provides another axis (i.e. number of particles) to scale performance with compute, with guaranteed gains. Recently, FK-steering in \cite{singhal2025general} was extended to video diffusion models with a reward function that governs the 3D/4D physical consistency~\citep{park2025steerx}.

%%%

\section{Extension to complex tasks}
\label{sec:extension}

\subsection{Blind inverse problems}
\label{subsec:blind}

Often, the forward operator $\Ac$ is parameterized with $\varphib$, i.e. $\Ac_\varphib$, unlike the problems that we have considered so far, which assumed full knowledge of the forward operator.
A prominent example is blind deconvolution, where the forward model is given as
\begin{align}
\label{eq:deblur}
    \y = \kb \ast \x + \n,
\end{align}
where $\kb$ is the convolution kernel. In such case, one has to specify the posterior of {\em both} $\x$ and $\kb$
\begin{align}
\label{eq:posterior_deblur}
    p(\x, \kb|\y) \propto p(\x)p(\kb)p(\y|\x, \kb),
\end{align}
where the factorization arises from the independence between $\x$ and $\kb$, and from \eqref{eq:deblur}, $p(\y|\x, \kb) = \Nc(\y, \kb \ast \xb, \sigma_y^2 I)$. In such case, $\kb = \varphib$.

\paragraph{BlindDPS~\citep{chung2023parallel}}

BlindDPS extends DPS by constructing another prior $p(\kb)$ for the kernel by training a separate diffusion model. Following the choice of \eqref{eq:pf_ode_simple}, one can construct two parallel PF-ODEs
\begin{align}
    d\x_t &= -t\nabla_{\x_t}\log p(\x_t)\,dt \\
    d\varphib_t &= -t\nabla_{\varphib_t}\log p(\varphib_t)\,dt
\end{align}
To be able to sample from the posterior given the measurement $\y$, we can create a coupling
\begin{align}
    d\x_t &= -t[\nabla_{\x_t}\log p(\x_t) + \nabla_{\x_t} \log p(\y|\x_t, \varphib_t)]\,dt \\
    d\varphib_t &= -t[\nabla_{\varphib_t} \log p(\varphib_t) + \nabla_{\varphib_t} \log p(\y|\x_t, \varphib_t)]\,dt.
\end{align}
Similar to the case of non-blind inverse problems, $p(\y|\x_t, \varphib_t)$ is intractable. BlindDPS uses the approximation proposed in DPS, but to both of the random variables $\x_t$ and $\varphib_t$, i.e. $p(\y|\x_t, \varphib_t) \approx p(\y|\hat\x_{0|t}, \varphib_{0|t})$, leading to
\begin{align}
\label{eq:blinddps_x}
    d\x_t &= -t[\nabla_{\x_t}\log p(\x_t) + \nabla_{\x_t} \log p(\y|\hat\x_{0|t}, \hat\varphib_{0|t})]\,dt \\
    d\varphib_t &= -t[\nabla_{\varphib_t} \log p(\varphib_t) + \nabla_{\varphib_t} \log p(\y|\hat\x_{0|t}, \hat\varphib_{0|t})]\,dt.
\label{eq:blinddps_k}
\end{align}
In practice, the BlindDPS requires sampling Gaussian noise independently for $\x_T$ and $\varphib_T$, then running \eqref{eq:blinddps_x} and \eqref{eq:blinddps_k} in parallel. The likelihood is approximated with the posterior mean of $\x_t$ and $\varphib_t$ at each step. Then, a gradient that maximizes this likelihood is applied separately to each stream.

\paragraph{GibbsDDRM~\citep{murata2023gibbsddrm}}
One downside of BlindDPS is that it requires training a score function for $\varphib$, and induces additional computational cost for calling $\nabla_{\varphib_t} \log p(\varphib_t)$ with a neural network. GibbsDDRM tackles the blind deblurring problem within the DDRM family. Formally, consider the SVD of $A$ that is dependent on the parameter of the forward operator $\varphib$, i.e. $A_\varphib = U_{\varphib}\Sigma_\varphib V_\varphib$, with singular values $\{s_{j,\bm{\varphi}}\}_{j=1}^m\}$. Similar to DDRM, let $\bar\y_\varphib := U_\varphib^\top \y_\varphib, \bar\x_\varphib := V_\varphib^\top \x_\varphib, \bar\epsilonb_\varphib := U_\varphib^\top \epsilonb_\varphib$, and further define $\bar\x_{0|t,\varphib} := V_\varphib \Ed[\x_0|\x_t]$. Then, the reverse distribution can be characterized as
\begin{align}
    p(\bar\vx_{t,\bm{\varphi}}^{(i)}|\vx_{t+1},\vy,\bm{\varphi}) = 
    \begin{cases}
        \gN(\bar\vx_{t,\bm{\varphi}}^{(i)}; \bar\vx_{0|t+1, \bm{\varphi}}^{(i)} + \sqrt{1 - \eta^2}\sigma_t\frac{\bar\vx_{t+1, \bm{\varphi}}^{(i)} - \bar\vx_{0|t+1, \bm{\varphi}}^{(i)}}{\sigma_{t+1}}, \eta^2\sigma_t^2) & \mbox{if  } s_{i,\bm{\varphi}} = 0\\
        \gN(\bar\vx_{t,\bm{\varphi}}^{(i)}; \bar\vx_{0|t+1, \bm{\varphi}}^{(i)} + \sqrt{1 - \eta^2}\sigma_t\frac{\bar\vy^{(i)}_{\bm{\varphi}} - \bar\vx_{0|t+1, \bm{\varphi}}^{(i)}}{\sigma_{\vy}/s_{i,\bm{\varphi}}}, \eta^2\sigma_t^2) & \mbox{if  } \sigma_t < \frac{\sigma_{\vy}}{s_{i,\bm{\varphi}}}\\
        \gN(\bar\vx_{t,\bm{\varphi}}^{(i)}; \bar\vy^{(i)}_{\bm{\varphi}}, \sigma_t^2 - \frac{\sigma_{\vy}^2}{s_{i,\bm{\varphi}}^2}) & \mbox{if  } \sigma_t \geq \frac{\sigma_{\vy}}{s_{i,\bm{\varphi}}} \\
    \end{cases}.
\label{eq:gibbsddrm}
\end{align}
Notice that \eqref{eq:gibbsddrm} assumes knowledge of $\varphib$. Akin to Gibbs sampling, the authors propose to update the random variable $\varphib$ with the following Langevin dynamics
\begin{align}
     \bm{\varphi} \leftarrow \bm{\varphi} + \frac{\xi}{2}\nabla_{\bm{\varphi}}\log p(\bm{\varphi}|\vx_{t:T},\vy) + \sqrt{\xi} \bm{\epsilon},
 \end{align}
 with some step size $\xi$. Following DPS, and placing the Laplacian prior on $\varphib$, the authors propose the following approximation
 \begin{align}
     \nabla_{\varphib} \log p(\varphib|\x_t,\yb) &= \nabla_\varphib \log p(\y|\x_{t:T}, \varphib) + \nabla_\varphib \log p(\varphib) \\
     &= \nabla_\varphib \log p(\y|\hat\x_{0|t}, \varphib) - \lambda \|\varphib\|_1,
 \end{align}
 with some constant $\lambda$.

 \paragraph{Fast Diffusion EM~\citep{laroche2024fast}}

Fast Diffusion EM takes an alternating expectation maximization (EM) approach. In the E-step, the approximated kernel $\varphib$ is used for the usual DPS/$\Pi$GDM sampling steps. In the M-step, an MAP optimization to maximize the posterior of the kernel is used, where the optimization problem is solved through a plug-and-play (PnP)~\citep{venkatakrishnan2013plug} method with a DnCNN~\citep{zhang2017beyond} denoiser. 

While the aforementioned approaches apply to a more general set of inverse problems, they are hard to apply to real-world image restoration tasks, as the forward model is either much more complicated, or hard to specify. For instance, the forward model of blind face restoration involves a convolution with a blur kernel, a down sampling operator, a noise component, and a JPEG degradation factor. \cite{man2025proxies} proposes to train a regressor to estimate these parameters, and show that using these estimated parameters together with an off-the-shelf inverse problem solver (e.g. DPS), is effective for solving inverse problems with complex forward operators.

\subsection{3D inverse problems}
\label{subsec:3d}

The inverse problems considered so far assume the latent signal $\x$ that we wish to retrieve is a 2D image. Due to architectural advances and the ease of data collection, it is fairly easy to collect a dataset of high-quality 2D images, and to train a diffusion model on it. Nevertheless, there are many cases in computational imaging, especially in biomedical imaging, where the reconstruction of 3D volume is necessary. In such cases, however, it is both hard to collect gold-standard 3D data, and to train a diffusion model on such collected 3D dataset. One popular way to tackle this is to decompose the prior
\begin{align}
\label{eq:factored_prior}
    p(\x) = \frac{1}{Z}\prod_{i=1}^K p_i(f_i(\x)),
\end{align}
where $Z$ is a normalization constant, and $f_i$ is an operator that captures complementary, lower-dimensional aspects of $\x$. A concrete example for the case of 3D would be to choose slicing operators for $f$, resulting in a factored prior over different {\em planes}.

\paragraph{DiffusionMBIR~\citep{chung2023solving}}

The core idea of DiffusionMBIR is that the 3D prior over $\x$ is already captured well in the 2D prior over the $xy$ slices. Thus, it may be sufficient to enforce smoothness across the other dimension, for instance, by using a total variation (TV) prior over the $z$ direction. This can be achieved by iteratively applying denoising steps and measurement consistency steps, where in the measurement consistency step, the following sub-problem is solved
\begin{align}
\label{eq:optimization_tv}
    \x' = \argmin_\x \frac{1}{2} \|\y - A\x\|_2^2 + \lambda\|D_z\x\|_1,
\end{align}
where $D_z$ is the finite difference operator across $z$. To solve \eqref{eq:optimization_tv}, ADMM~\citep{boyd2011distributed} is used, with CG steps operating to solve the inner problem. However, notice that this would require immense computation cost, as the iterative ADMM would have to be solved for every $t$. To mitigate this cost, a variable sharing technique was proposed so that the primal and dual variables are warm-started from the previous iteration $t+1$, and only a single iteration of ADMM is applied to each optimization step. Later, in \cite{chung2024decomposed}, it was shown that one can improve the performance of DiffusionMBIR by using the Tweedie estimates $\hat\x_{0|t}$ for optimization in \eqref{eq:optimization_tv}, instead of the noisy variables $\x_t$.

\paragraph{TPDM~\citep{lee2023improving}}

Another way to construct a factored prior is to use two diffusion diffusion priors for different slice directions. Compared to DiffusionMBIR, this further alleviates hand-crafted inductive bias and replaces it with a data-driven generative prior, and was shown to outperform DiffusionMBIR across several tasks, especially on tasks such as super-resolution. One way to implement the product distribution is by using the sum of the scores
\begin{align}
\label{eq:tpdm_sum}
    \nabla_{\x_t} \log p(\x_t) = \alpha \nabla_{\x_t} \log q^{(p)}(\x_t) + \beta \nabla_{\x_t} \log q^{(a)}(\x_t), 
\end{align}
where $q^{(p)}$ is the distribution of the slices in the primary plane, $q^{(a)}$ is the distribution of the slices in the auxiliary plane (i.e. orthogonal to the primary plane), and $\alpha,\beta$ are mixing constants. In practice, directly using \eqref{eq:tpdm_sum} would incur double the computation cost during inference. To mitigate this, \cite{lee2023improving} proposes to use an alternating approach, using only the score from the primary plane for $\frac{\alpha}{\alpha+\beta}$ fraction of the time during reverse sampling, and using only the score from the auxiliary plane for $\frac{\beta}{\alpha+\beta}$ for the rest. In order to impose measurement consistency, DPS steps are employed.

\subsection{Inverse problems under data scarcity}
\label{subsec:ood}

All diffusion model-based inverse problem solvers rely on the assumption that one has access to a diffusion model trained on high-quality in-distribution datasets. This condition is not satisfied. For instance, in black-hole imaging~\citep{akiyama2019first} and cryo-EM imaging~\citep{gupta2021cryogan}, one only has access to the partial measurements, with no access whatsoever on how the {\em true} image would look like. In this section, we review some of the approaches that operate under such constraints.

\subsubsection{Test-time adaptation}
\label{subsubsec:adaptation}

One way to solve this problem is to use a diffusion model trained on a separate dataset, and try to adapt the diffusion model on out-of-distribution (OOD) measurements online~\cite{barbano2025steerable,chung2024deep}. The approaches build on top of deep image prior (DIP)~\citep{ulyanov2018deep}, which overfits a network on a single measurement, relying on the inductive prior of the neural network
\begin{align}
\label{eq:dip}
    \theta^* = \argmin_\theta \|\y - AG_\theta(\z)\|_2^2,
\end{align}
where $G_\theta$ is the network for reconstruction, which takes in a random input $\z \sim \Nc(0, I)$.

Deep Diffusion Image Prior (DDIP)~\citep{chung2024deep} generalizes and extends DIP to work within the diffusion framework by alternating the following steps
\begin{align}
\label{eq:ddip}
    \text{for } t = T, \ldots, 1:
    \theta_{t-1} \gets &\argmin_{\theta_t} \|\y - \Ab D_{\theta_t}(\x_t|\y)\|_2^2, \\
    \x_{t-1} \gets &\ddim_{\theta_{t-1}}(D_{\theta_{t-1}}(\x_t|\y), \eta),
\label{eq:ddip2}
\end{align}
where $\ddim_\theta(\x_t,\eta) := \sqrt{\bar\alpha_{t-1}}D_\theta(\x_t|\y) + \sqrt{1 - \bar\alpha_{t-1}}\left(\eta\epsilonb + (1 - \eta)\epsilonb^\theta\right)$. Notice that DDIP differs from DIP in two aspects. First, the reconstructor $G_\theta$ is replaced with an MMSE denoiser $D_\theta$, which stems from a pre-trained diffusion model, and hence the generation trajectory is pivoted in the original generative process. Second, the DIP adaptation in \eqref{eq:ddip} is held across multiple scales (i.e. noise levels $t$), different from a single-scale optimization of DIP. In practice, the original parameters $\theta$ are hold constant, and only the low-rank adaptation (LoRA) is applied to make partial updates to the network.

\paragraph{Patch-based priors}
Factored priors that are widely employed within the 3D medical imaging setting, but were also shown to be useful for 2D inverse problems, for instance, by using patch-based priors~\citep{hu2024learning}. By employing positional encodings, PaDIS~\citep{hu2024learning} constructs a position-aware patch-based diffusion model, showing that such approach is better than image diffusion model counterparts, especially in the data-scare regime. Later, the patch-based diffusion approach was combined with test-time adaptation in \cite{hu2025test}.

\subsection{Training a diffusion model with noisy data}

\paragraph{GSURE-based diffusion model~\citep{kawar2024gsurebased}}

Stein's Unbiased Risk Estimator (SURE)~\citep{stein1981estimation} is a widely used method to train a denoiser given only the Gaussian-noisy measurements. Later, this was extended to a general set of linear inverse problems of the form \eqref{eq:ip_base} in Generalized SURE (GSURE)~\citep{eldar2008generalized}, which states the following
\begin{align}
\label{eq:gsure}
    \Ed\left[
    \|P(D(\y) - \x)\|_2^2
    \right] = \Ed\left[
    \|P(D(\y) - \x_{ML}\|_2^2
    \right] + 2\Ed\left[
    \nabla_{A^\top \y} \cdot PD(y)
    \right] + c,
\end{align}
where $P = A^\top A$ and $\x_{ML} = (A^\top A)^\dagger A^\top \y$. While \eqref{eq:gsure} guarantees a good denoiser in the sense of projected MSE, this ceases to be a good surrogate when the operator $A$ removes sufficient information from $\x$ (i.e. when the mask is large). In such case, one can use ENsmeble SURE (ENSURE)~\citep{aggarwal2022ensure} by also marginalizing over the operator $A$, given the assumption that we have access to $A$ and the noise level, and the different realizations of $A$ covers the signal space $\Rd^n$. Note that this assumption is satisfied, for instance, in MRI acquisitions.
GSURE-based diffusion follows this assumption and leverages ENSURE to train a diffusion model from the measurements only.

Following the similar procedure from the DDRM family introduced in Sec.~\ref{subsec:DDRM family}, we transform the inverse problem into
\begin{align}
    \bar\y = P\bar\x + \bar\z, \quad \bar\z \sim \Nc(0, \sigma_y^2 \Sigma^\dagger \Sigma^{\dagger\top}),
\end{align}
where $A = U\Sigma V^\top, P = \Sigma^\dagger \Sigma, \bar\x = V^\top \x, \bar\y = \Sigma^\dagger U^\top \y, \bar\z = \Sigma^\dagger U^\top \z$. GSURE-diffusion then constructs the following forward perturbation
\begin{align}
    \bar\x_t = \sqrt{\bar\alpha_t}\y + \left(
    (1 - \bar\alpha_t)I - \bar\alpha_t \sigma_y^2 \Sigma^\dagger \Sigma^{\dagger\top}
    \right)^{1/2} \epsilonb,
\end{align}
and by this design choice, the marginal distribution of $\bar\x_t$ reads $q(\bar\x_t|\bar\x, P) = \Nc(\sqrt{\bar\alpha_t}P\bar\x, (1 - \bar\alpha_t)I)$. The objective function then reads
\begin{align}
\label{eq:gsure_based_diffusion}
    \sum_{t=1}^T \gamma_t \Ed\left[
    \left\|WP\left(
    D_\theta(\bar\x_t) - \frac{1}{\sqrt{\bar\alpha_t}}\bar\x_t
    \right)\right\|_2^2 + 2(1 - \bar\alpha_t)\left(
    \nabla_{\bar\x_t} \cdot PW^2 D_\theta(\bar\x_t)
    \right) + c
    \right],
\end{align}
where $W = \Ed[P]^{-\frac{1}{2}} \succ 0$. It was shown in \cite{kawar2024gsurebased} that by training a diffusion model solely on measuremenets with \eqref{eq:gsure_based_diffusion} yields similar to performance to the diffusion models trained on clean samples $\x$.

\subsubsection{Ambient Diffusion Family}

\paragraph{Ambient Diffusion~\citep{daras2023ambient}}

Ambient Diffusion considers a special case of learning a diffusion model from noiseless-masked measurements $\y_0 = A\x_0$ with the same assumptions as in GSURE-diffusion. Consider the following naive loss
\begin{align}
\label{eq:ambient_diffusion_naive}
    J^{\rm naive}(\theta) = \Ed\left[
    \left\|
    A(D_\theta(A, A\x_t, t) - \x_0)
    \right\|_2^2
    \right],
\end{align}
where the loss simply ignores the missing pixels, and computes the loss only on known ones. Training a diffusion model with \eqref{eq:ambient_diffusion_naive}, however, would not lead the network to learn any information about the {\em unknown} pixel values. To mitigate this, the authors propose to sample a second mask $B$, and set $\tilde{A} = BA$. Then, the loss of Ambient Diffusion reads
\begin{align}
\label{eq:ambient_diffusion}
    J^{\rm corr}(\theta) = \Ed\left[
    \left\|
    A(D_\theta(\tilde{A}, \tilde{A}\x_t, t) - \x_0)
    \right\|_2^2
    \right]
\end{align}
Since the network $D_\theta$ cannot distinguish between the old and new masked pixels, the safest way would be to reconstruct every pixel. Under mild assumptions on $A,B$, one can also show that $D_{\theta^*}(\tilde{A}, \x_t) = \Ed[\x_0|\tilde{A}\x_t, \tilde{A}]$.

\paragraph{Consistent Diffusion meets Tweedie~\citep{daras2024consistent}}

\cite{daras2024consistent} considers training a diffusion model with Gaussian noise-corrupted samples, where $A = I$. Let the noise level of the samples be $t_n$. Notice that for $t > t_n$, we can express the random variable $\x_t$ in two distinct ways: $\x_t = \x_0 + \sigma_t \epsilonb$ and $\x_t = \x_{t_n} + \sqrt{\sigma_t^2 - \sigma_{t_n}^2}\epsilonb$. By applying Tweedie's formula twice, one can conclude
\begin{align}
\label{eq:tweedie_twice}
    \Ed[\x_{t_n}|\x_t] = \frac{\sigma_t^2 - \sigma_{t_n}^2}{\sigma_t^2}\left(
    \Ed[\x_0|\x_t] - \x_t
    \right) + \x_t.
\end{align}
The implication of \eqref{eq:tweedie_twice} is that one can train an optimal denoiser for noise levels $t > t_n$ by training the model to remove only the additional noise from $t_n$ to $t$, i.e. train the model with
\begin{align}
\label{eq:objective_tweedie_twice}
    \Ed_{t \sim \Uc(t_n, T]}\left[\left\|
    \frac{\sigma_t^2 - \sigma_{t_n}^2}{\sigma_t^2}D_\theta(\x_t, t) + \frac{\sigma_{t_n}^2}{\sigma_t^2}\x_t - \x_{t_n}
    \right\|_2^2\right].
\end{align}
For noise levels $t < t_n$, one can leverage the idea from Consistent diffusion~\citep{daras2023consistent}, where the objective reads
\begin{align}
\label{eq:objective_consistent_diffusion}
    \Ed_{t \sim \Uc(t_n, T], t' \sim \Uc(\varepsilon, T), t'' \sim \Uc(t' - \varepsilon, t')}\left[\left\|
    D_\theta(\x_{t'}, t') - \Ed_{\x_{t''} \sim p_\theta(\x_{t''}|\x_{t'})} [D_\theta(\x_{t''}, t'')]
    \right\|_2^2\right].
\end{align}
Notice that in \eqref{eq:objective_consistent_diffusion}, $t' > t''$, and the sampling of $p_\theta(\x_{t''}|\x_{t'})$ is achieved through taking a step $\varepsilon$ through a diffusion model, a similar procedure to consistency models~\citep{song2023consistency}. By training a diffusion model to be consistent with its counterpart taken two steps, one can achieve an optimal denoiser even for $t < t_n$. Thus, the final objective of \cite{daras2024consistent} takes a weighted sum of the two objectives \eqref{eq:objective_tweedie_twice}, \eqref{eq:objective_consistent_diffusion}.

\paragraph{Ambient Diffusion Omni~\citep{daras2025ambient}}

Previous works relied on the assumption that one knows the noise level of the corrupted measurement. Ambient Diffusion Omni relaxes this assumption and considers the case where the diffusion model is trained on a mixture distribution of $p_0$ and $q_0$, where $p_0$ is the clean data distribution, and $q_0$ is the corrupted distribution containing arbitrary mix of bad-quality data (e.g. blur, noise, JPEG artifacts, etc.). In the practical scenario when training a diffusion model for deployment, one would filter out the samples in $q_0$ and use only the ones in $p_0$. However, in Ambient Diffusion Omni, the authors propose a way to utilize both the data from $p_0$ and $q_0$, showing that one can achieve {\em better} quality by using data from both sources.

Due to the contracting property of diffusion models~\citep{chung2022come}, when noise is added, $p_t$ and $q_t$ become closer to each other. The key idea of Ambient Diffusion Omni is to train a classifier that distinguishes high and low quality samples at a certain timestep. The minimum timestep $t_n^{\rm min}$ is distinguished for each sample. Then, when training the diffusion model, one only uses the timesteps $t \geq t_n^{\rm min}$ for some sample $n$.

\subsubsection{Expectation-Maximization (EM)}

EM tries to find the best parameter $\theta$ of the model that best explains the observation $\y$. The challenge is that we do not know the underlying clean data $\x$. To circumvent this issue, EM takes a two-stage approach.

\begin{enumerate}
    \item \textbf{(E-step)}: Use the current model $\theta_k$ to specify the posterior $p_{\theta_k}(\x|\y)$ and specify {\em in expectation} what the complete data looks like
    \item \textbf{(M-step)}: Given the probabilistic guess about the hidden data $\x$, maximize the log-likelihood of the model to get $\theta_{k+1}$
\end{enumerate}

The key idea behind the EM algorithm is that for any $\theta_a$ and $\theta_b$, we have
\begin{align}
    \log\frac{p_{\theta_a}(\x)}{p_{\theta_b}(\x)} \geq \Ed_{p(\y)}\Ed_{q_{\theta_b}(\x|\y)}\left[\frac{p_{\theta_a}(\x)}{p_{\theta_b}(\x)}\right]
\end{align}
Hence, the iteration of EM leads to a sequence of parameters $\theta_k$ where the expected log evidence $\Ed_{p(\y)}[\log p_{\theta_k}(\y)]$ monotonically increases and converges to a local optimum.

\paragraph{\cite{rozet2024learning}}

Notice that
\begin{align}
    \theta_{k+1} &= \argmax_\theta \Ed_{p(\y)}\Ed_{p_{\theta_k}(\x|\y)}[\log p_\theta(\x) + \log p(\y|\x)] \\
    &= \argmax_\theta \Ed_{p(\y)}\Ed_{p_{\theta_k}(\x|\y)}[\log p_\theta(\x)] \\
    &= \argmin_\theta {\rm KL}(\pi_k(\x) \| p_\theta(\x)),
\end{align}
where $\pi_k(\x) = \int p_{\theta_k}(\x|\y) p(\y)\,d\y$. In practice, given a sample $\y$, the authors propose to use a posterior sampler, namely the moment matching method discussed in Sec.~\ref{sec:explicit} to draw from $\pi_k(\x)$. Then, with the collected samples, the M-step is performed by standard DSM.
A concurrent work of \cite{bai2024expectation} uses the same EM framework, but uses DPS to draw posterior samples in the E-step.

\section{Text-driven solutions}
\label{sec:text}

Since inverse problems are ill-posed, measurement does not provide sufficient information for perfect recovery. It is natural that if one could use additional auxiliary information for recovery, it would be beneficial to do so. As one such side information, text has recently gained attention, as they enable compact, informative, and highly versatile conditioning.

Often, Latent Diffusion Models (LDMs)~\citep{vahdat2021score,rombach2022high}  enable effective entanglement of multi-modal representations, and are considered the de facto standard for modern text-to-image diffusion models.
In this section, let $\vz_0 = \Ec(\vx_0)$ be a latent code of clean image encoded by VAE encoder $\Ec$. Original image can be reconstructed $\vx_0 = \Dc(\vz_0)=\Dc(\Ec(\vx_0))$ by VAE decoder $\Dc$. The diffusion model is now defined on the latent space.

\paragraph{P2L (\cite{chung2024prompttuning})}

P2L demonstrates the effectiveness of text embedding space for improving quality of solution. 
The authors propose an extension of DPS for LDMs~\citep{rout2024solving}, by using
\begin{align}
    \nabla_{\vz_t} \log p(\vy|\vz_t) \approx \nabla_{\vz_t} \log p(\vy|\Dc(\mathbb{E}[\vz_0|\vz_t])).
\end{align}
Now, to consider the text embedding as an optimization variable, they formulate an optimization problem 
\begin{align}
    \min_{\vz\sim p(\vz|\vy)} \min_\vc \quad &\| \vy - A\Dc(\vz^{(\vc)}) \| ^2\\
    \text{subject to} \quad &\vz \in F_X\nonumber
\end{align}
where $F_X = \{\vz| \z=\Ec(\vx) \text{for some} \vx\}$ denotes the set of latent that can be represented by some image $\vx$.
Optimization involves two alternative updates,
\begin{align}
    \vc^* = \argmin_\vc \|\vy - A\Dc(\hat\vz_{0|t}^{(\vc)}) \|^2
\end{align}
for the prompt embedding and
\begin{align}
    \vz^* &= \Ec(\vx^*) \quad \text{where}\nonumber\\
    \vx^* &= \argmin_\vx \|\vy-A\vx\|^2 + \lambda \|\vx - \Dc(\hat\vz_{0|t}^{(c^*)}) \|^2
\end{align}
for the MMSE estimate during reverse sampling,
which considers not only data fidelity but also latent fidelity, encouraging the optimized latent to lie within the range of VAE encoder. This ensures that the decoded output remains on the image manifold.
The joint update of prompt embedding during diffusion reverse sampling leads solution to be more aligned to the pre-trained diffusion prior, compared to using null-text embedding.

\paragraph{TReg (\cite{kim2025regularization})}
While P2L reduces the gap between the latent diffusion prior and the solution obtained via null-text embedding, it does not leverage the text prompt as an additional prior to guide the solution.
To address this limitation, TReg introduces the concept of \textit{Regularization by text}, which further constrains the solution space toward a conditional prior distribution, implemented via a latent-space optimization problem.
By applying Bayes' rule to posterior distribution involving latent variable $\vz$, we obtain:
\begin{align}
    p(\vx|\vy, \vz)\propto p(\vy|\vx,\vz)p(\vx|\vz) \propto p(\vz|\vx,\vy)p(\vy|\vx)p(\vx|\vz).
\end{align}
TReg formulates a Maximum A Posterior (MAP) optimization problem with text regularization term applied on MMSE estimate space during reverse sampling as
\begin{align}
    \min_{\vz, \vx} \underbrace{\|\vz-\Ec(\Dc(\vz))\|^2 + \|\vy-A\Dc(\vz)\|^2}_{\ell_{\text{MAP}}} + \lambda \underbrace{\| \vz - \hat\vz_{0|t}\|^2}_{ \ell_{\text{Treg}}}\quad \text{s.t.} \quad \vx = \Dc(\vz)
\end{align}
where $p(\vx|\vz):=\delta(\vx-\Dc(\vz))$, $\hat\vz_{0|t}$ denotes text conditioned denoised estimate, and $\vz$ is initialized with $\hat\vz_{0|t}$. The regularization term steers the sampling trajectory toward a clean manifold aligned with the text condition $\vc$. Combined with the MAP objective enforcing data fidelity, this approach yields solutions that satisfies both the text condition and data consistency with given measurement, thereby improving reconstruction quality, especially under severe degraded conditions.
TReg also introduces adaptive negation, which optimizes the null-text embedding to suppress concepts unrelated to the text condition $\vc$ by minimizing CLIP similarity with the denoised estimate.

\paragraph{ContextMRI~\citep{chung2025contextmri}}

Text-driven solutions were also adopted in the medical imaging domain, by using metadata as a conditioning signal. The leveraged metadata include patient demographics, the location including slice number and anatomy, MRI imaging parameters including TR, TE, TI, and even (optionally) pathology. The authors trained a diffusion model in the pixel space with a CLIP encoder that takes in as input the metadata represented as text, and use this model as the prior for MRI reconstruction. It was shown that in all cases, the conditional diffusion model performs better than the unconditional counterpart.

\section{Discussion and Conclusion}
\label{sec:discussion_and_conclusion}

In this chapter, we gave a comprehensive overview of using diffusion models for inverse problems, primarily focused on the general zero-shot solvers that does not involve task-specific training, and thus can be adapted to various applications. Our survey deliberately omitted two related areas to maintain this focus: solvers designed explicitly for Latent Diffusion Models (LDMs)~\citep{rout2023solving,rout2024beyond,raphaeli2025silo}—whose underlying principles largely align with the methods discussed—and approaches based on diffusion bridges~\citep{delbracio2023inversion,luo2023image,liu2023i2sb,chung2023direct}, which necessitate supervised training.

A key takeaway is the inherent trade-off among the surveyed methods. Solvers present a spectrum of design choices, balancing computational speed against reconstruction fidelity and exactness. The selection of an appropriate method is therefore contingent upon the specific constraints and goals of the target application.

The diversity of these powerful techniques signifies a rapidly maturing field. As these tools become more robust, they offer practitioners a versatile and adaptable toolkit for a wide range of scientific and creative applications. Future work will likely focus on reconciling the trade-offs between speed and accuracy, pushing the boundaries of what is achievable in unsupervised inverse problem-solving.

%%%

{
    \bibliographystyle{agsm}
    \bibliography{main.bib}
}

\end{document}

%% file: packages.tex
\usepackage{natbib}
\usepackage[utf8]{inputenc} % allow utf-8 input
\usepackage[T1]{fontenc}    % use 8-bit T1 fonts
\usepackage{url}            % simple URL typesetting
\usepackage{booktabs}       % professional-quality tables
\usepackage{amsfonts}       % blackboard math symbols
\usepackage{nicefrac}       % compact symbols for 1/2, etc.
\usepackage{microtype}      % microtypography
\usepackage{xcolor}       % colors
\usepackage{xkcdcolors}
\usepackage{times}
\usepackage{epsfig}
\usepackage{graphicx}
\usepackage{placeins}
\usepackage{multirow}
\usepackage[skip=5pt]{caption}
\usepackage{rotating}
\usepackage{cancel}
\usepackage{setspace}
\usepackage{eso-pic}

\usepackage{bm}
\usepackage{bibunits} % for separate ref list in appendix

\usepackage{amssymb,amsthm}
\usepackage[hidelinks]{hyperref}
\usepackage{amsmath}
\usepackage{graphicx}
\usepackage{wrapfig,lipsum,booktabs}

\usepackage{epsfig}
\usepackage{tikz}
\usetikzlibrary{spy}
\usepackage{algpseudocode}
\usepackage{algorithm}
\usepackage{mathrsfs}

\usepackage{nicefrac}       % compact symbols for 1/2, etc.
\usepackage{booktabs}       % professional-quality tables

\usepackage{thmtools,thm-restate}
\usepackage{cleveref}
\usepackage{listings}
\usepackage{lstautogobble}  %
\usepackage{color}          %
\usepackage{zi4}            %
\definecolor{bluekeywords}{rgb}{0.13, 0.13, 1}
\definecolor{greencomments}{rgb}{0, 0.5, 0}
\definecolor{redstrings}{rgb}{0.9, 0, 0}
\definecolor{graynumbers}{rgb}{0.5, 0.5, 0.5}
\usepackage{subcaption}        % subfigure
\usepackage{siunitx}               % <--- new
\usepackage[export]{adjustbox}
\usepackage{makecell}
\usepackage{url}
\usepackage{tikz, pgfplots}
\usetikzlibrary{positioning}
\usepackage{capt-of}
\usepackage{diagbox}

\def\eqref#1{(\ref{#1})}

\usepackage{colortbl}

\usepackage{mdframed}

%% file: macros.tex
\usepackage{amsmath,amsfonts,bm}

\def\eqref#1{Eq.~(\ref{#1})}
\def\1{\bm{1}}

\def\rmI{{\mathbf{I}}}

\newcommand{\f}{{\boldsymbol f}}

\def\vmu{{\bm{\mu}}}

\def\vb{{\bm{b}}}
\def\vc{{\bm{c}}}

\def\vs{{\bm{s}}}

\def\vx{{\bm{x}}}
\def\vy{{\bm{y}}}
\def\vz{{\bm{z}}}

\newcommand{\Ib}{{\bm I}}

\DeclareMathAlphabet{\mathsfit}{\encodingdefault}{\sfdefault}{m}{sl}
\SetMathAlphabet{\mathsfit}{bold}{\encodingdefault}{\sfdefault}{bx}{n}

\def\gN{{\mathcal{N}}}

\DeclareMathOperator*{\argmax}{arg\,max}
\DeclareMathOperator*{\argmin}{arg\,min}

\newcommand{\xb}{{\boldsymbol x}}
\newcommand{\yb}{{\boldsymbol y}}

\newcommand{\kb}{{\boldsymbol k}}

\newcommand{\s}{{\boldsymbol s}}
\newcommand{\n}{{\boldsymbol n}}
\newcommand{\x}{{\boldsymbol x}}
\newcommand{\y}{{\boldsymbol y}}
\newcommand{\z}{{\boldsymbol z}}
\newcommand{\w}{{\boldsymbol w}}

\newcommand{\epsilonb}{{\boldsymbol \epsilon}}
\newcommand{\mub}{{\boldsymbol \mu}}

\newcommand{\Ab}{{\boldsymbol A}}

\newcommand{\varphib}{{\bm \varphi}}

\newcommand{\Ed}{{\mathbb E}}
\newcommand{\Rd}{{\mathbb R}}

\newcommand{\Ac}{{\mathcal A}}

\newcommand{\Dc}{{\mathcal D}}
\newcommand{\Ec}{{\mathcal E}}

\newcommand{\Nc}{{\mathcal N}}
\newcommand{\Mc}{{\mathcal M}}

\newcommand{\Pc}{{\mathcal P}}

\newcommand{\Uc}{{\mathcal U}}

\usetikzlibrary{positioning}
\usepackage{capt-of}
\usepackage{diagbox}

\definecolor{C0}{rgb}{0.121569, 0.466667, 0.705882}
\definecolor{C1}{rgb}{1.000000, 0.498039, 0.054902}
\definecolor{C2}{rgb}{0.172549, 0.627451, 0.172549}
\definecolor{C3}{rgb}{0.839216, 0.152941, 0.156863}
\definecolor{C4}{rgb}{0.580392, 0.403922, 0.741176}
\definecolor{C5}{rgb}{0.549020, 0.337255, 0.294118}
\definecolor{C6}{rgb}{0.890196, 0.466667, 0.760784}
\definecolor{C7}{rgb}{0.498039, 0.498039, 0.498039}
\definecolor{C8}{rgb}{0.737255, 0.741176, 0.133333}
\definecolor{C9}{rgb}{0.090196, 0.745098, 0.811765}
\definecolor{trolleygrey}{rgb}{0.5, 0.5, 0.5}

\definecolor{BrickRed}{rgb}{0.6,0,0}
\definecolor{RoyalBlue}{rgb}{0,0,0.8}
\definecolor{Tdgreen}{rgb}{0,0.4,0.7}
\definecolor{pinegreen}{rgb}{0.0, 0.47, 0.44}
\definecolor{cornellred}{rgb}{0.7, 0.11, 0.11}
\definecolor{cadmiumgreen}{rgb}{0.0, 0.42, 0.24}
\definecolor{spirodiscoball}{rgb}{0.06, 0.75, 0.99}
\definecolor{mylightblue}{rgb}{0.85, 0.90, 0.94}
\definecolor{maroon}{cmyk}{0,0.87,0.68,0.32}

\usepackage{pifont}% http://ctan.org/pkg/pifont
\algnewcommand{\LineComment}[1]{\State \(\triangleright\) #1}

\definecolor{cfg}{rgb}{0.906, 0.435, 0.318}
\definecolor{cfgpp}{rgb}{0.165, 0.616, 0.561}
\definecolor{cfgnull}{rgb}{0.208, 0.565, 0.953}

\newcommand{\ddim}{{\rm DDIM}}

\def\eqref#1{(\ref{#1})}
\usepackage{tikz}

\definecolor{cornellred}{rgb}{0.7, 0.11, 0.11}
\definecolor{cadmiumgreen}{rgb}{0.0, 0.42, 0.24}
\definecolor{aliceblue}{rgb}{0.91, 0.94, 0.97}
\definecolor{darkblue}{rgb}{0.83, 0.89, 0.97}
\definecolor{Red7}{rgb}{0.941, 0.243, 0.243}
\definecolor{Green7}{RGB}{55, 178, 77}
\definecolor{Blue9}{rgb}{0.098,0.3,0.9}

\usepackage{pifont}% http://ctan.org/pkg/pifont

\usepackage{soul}
\sethlcolor{aliceblue}

\hypersetup{
  linkcolor = cornellred,
  citecolor  = cadmiumgreen,
  colorlinks = true,
  urlcolor = Blue9
}

%% file: main.bib
@article{ho2020denoising,
  title={Denoising diffusion probabilistic models},
  author={Ho, Jonathan and Jain, Ajay and Abbeel, Pieter},
  journal={Advances in Neural Information Processing Systems},
  volume={33},
  pages={6840--6851},
  year={2020}
}

@inproceedings{
    chung2024decomposed,
    title={Decomposed Diffusion Sampler for Accelerating Large-Scale Inverse Problems},
    author={Hyungjin Chung and Suhyeon Lee and Jong Chul Ye},
    booktitle={The Twelfth International Conference on Learning Representations},
    year={2024},
    url={https://openreview.net/forum?id=DsEhqQtfAG}
}

@inproceedings{sohl2015deep,
  title={Deep unsupervised learning using nonequilibrium thermodynamics},
  author={Sohl-Dickstein, Jascha and Weiss, Eric and Maheswaranathan, Niru and Ganguli, Surya},
  booktitle={International Conference on Machine Learning},
  pages={2256--2265},
  year={2015},
  organization={PMLR}
}

@inproceedings{rombach2022high,
  title={High-resolution image synthesis with latent diffusion models},
  author={Rombach, Robin and Blattmann, Andreas and Lorenz, Dominik and Esser, Patrick and Ommer, Bj{\"o}rn},
  booktitle={Proceedings of the IEEE/CVF Conference on Computer Vision and Pattern Recognition},
  pages={10684--10695},
  year={2022}
}

@inproceedings{
kawar2022denoising,
title={Denoising Diffusion Restoration Models},
author={Bahjat Kawar and Michael Elad and Stefano Ermon and Jiaming Song},
booktitle={Advances in Neural Information Processing Systems},
editor={Alice H. Oh and Alekh Agarwal and Danielle Belgrave and Kyunghyun Cho},
year={2022},
url={https://openreview.net/forum?id=kxXvopt9pWK}
}

@inproceedings{
chung2022improving,
title={Improving Diffusion Models for Inverse Problems using Manifold Constraints},
author={Hyungjin Chung and Byeongsu Sim and Dohoon Ryu and Jong Chul Ye},
booktitle={Advances in Neural Information Processing Systems},
editor={Alice H. Oh and Alekh Agarwal and Danielle Belgrave and Kyunghyun Cho},
year={2022},
url={https://openreview.net/forum?id=nJJjv0JDJju}
}

@inproceedings{
chung2023diffusion,
title={Diffusion Posterior Sampling for General Noisy Inverse Problems},
author={Hyungjin Chung and Jeongsol Kim and Michael Thompson Mccann and Marc Louis Klasky and Jong Chul Ye},
booktitle={International Conference on Learning Representations},
year={2023},
url={https://openreview.net/forum?id=OnD9zGAGT0k}
}

@article{chung2023solving,
  title={Solving 3D Inverse Problems using Pre-trained 2D Diffusion Models},
  author={Chung, Hyungjin and Ryu, Dohoon and Mccann, Michael T and Klasky, Marc L and Ye, Jong Chul},
  journal={IEEE/CVF Conference on Computer Vision and Pattern Recognition},
  year={2023}
}

@article{chung2023parallel,
  title={Parallel Diffusion Models of Operator and Image for Blind Inverse Problems},
  author={Chung, Hyungjin and Kim, Jeongsol and Kim, Sehui and Ye, Jong Chul},
  journal={IEEE/CVF Conference on Computer Vision and Pattern Recognition},
  year={2023}
}

@article{vincent2011connection,
  title={A connection between score matching and denoising autoencoders},
  author={Vincent, Pascal},
  journal={Neural computation},
  volume={23},
  number={7},
  pages={1661--1674},
  year={2011},
  publisher={MIT Press}
}

@inproceedings{song2019generative,
 author = {Song, Yang and Ermon, Stefano},
 booktitle = {Advances in Neural Information Processing Systems},
 pages = {},
 title = {Generative Modeling by Estimating Gradients of the Data Distribution},
 volume = {32},
 year = {2019}
}

@inproceedings{song2020score,
  author    = {Yang Song and
               Jascha Sohl{-}Dickstein and
               Diederik P. Kingma and
               Abhishek Kumar and
               Stefano Ermon and
               Ben Poole},
  title     = {Score-Based Generative Modeling through Stochastic Differential Equations},
  booktitle = {9th International Conference on Learning Representations, {ICLR}},
  year      = {2021},
}

@article{huang2021variational,
  title={A Variational Perspective on Diffusion-Based Generative Models and Score Matching},
  author={Huang, Chin-Wei and Lim, Jae Hyun and Courville, Aaron},
  journal={arXiv preprint arXiv:2106.02808},
  year={2021}
}

@inproceedings{bora2017compressed,
  title={Compressed sensing using generative models},
  author={Bora, Ashish and Jalal, Ajil and Price, Eric and Dimakis, Alexandros G},
  booktitle={International conference on machine learning},
  pages={537--546},
  year={2017},
  organization={PMLR}
}

@inproceedings{lee2023improving,
  title={Improving 3D imaging with pre-trained perpendicular 2D diffusion models},
  author={Lee, Suhyeon and Chung, Hyungjin and Park, Minyoung and Park, Jonghyuk and Ryu, Wi-Sun and Ye, Jong Chul},
  booktitle={Proceedings of the IEEE/CVF International Conference on Computer Vision},
  pages={10710--10720},
  year={2023}
}

@inproceedings{
trippe2023diffusion,
title={Diffusion Probabilistic Modeling of Protein Backbones in 3D for the motif-scaffolding problem},
author={Brian L. Trippe and Jason Yim and Doug Tischer and David Baker and Tamara Broderick and Regina Barzilay and Tommi S. Jaakkola},
booktitle={The Eleventh International Conference on Learning Representations },
year={2023},
url={https://openreview.net/forum?id=6TxBxqNME1Y}
}

@article{park2025steerx,
  title={SteerX: Creating Any Camera-Free 3D and 4D Scenes with Geometric Steering},
  author={Park, Byeongjun and Go, Hyojun and Nam, Hyelin and Kim, Byung-Hoon and Chung, Hyungjin and Kim, Changick},
  journal={arXiv preprint arXiv:2503.12024},
  year={2025}
}

@article{singhal2025general,
  title={A general framework for inference-time scaling and steering of diffusion models},
  author={Singhal, Raghav and Horvitz, Zachary and Teehan, Ryan and Ren, Mengye and Yu, Zhou and McKeown, Kathleen and Ranganath, Rajesh},
  journal={arXiv preprint arXiv:2501.06848},
  year={2025}
}

@inproceedings{
cardoso2024monte,
title={Monte Carlo guided Denoising Diffusion models for Bayesian linear inverse problems.},
author={Gabriel Cardoso and Yazid Janati el idrissi and Sylvain Le Corff and Eric Moulines},
booktitle={The Twelfth International Conference on Learning Representations},
year={2024},
url={https://openreview.net/forum?id=nHESwXvxWK}
}

@inproceedings{
dou2024diffusion,
title={Diffusion Posterior Sampling for Linear Inverse Problem Solving: A Filtering Perspective},
author={Zehao Dou and Yang Song},
booktitle={The Twelfth International Conference on Learning Representations},
year={2024},
url={https://openreview.net/forum?id=tplXNcHZs1}
}

@article{song2021maximum,
  title={Maximum likelihood training of score-based diffusion models},
  author={Song, Yang and Durkan, Conor and Murray, Iain and Ermon, Stefano},
  journal={Advances in Neural Information Processing Systems},
  volume={34},
  year={2021}
}

@inproceedings{chen2018neural,
 author = {Chen, Ricky T. Q. and Rubanova, Yulia and Bettencourt, Jesse and Duvenaud, David K},
 booktitle = {Advances in Neural Information Processing Systems},
 pages = {},
 title = {Neural Ordinary Differential Equations},
 volume = {31},
 year = {2018}
}

@article{hyvarinen2005estimation,
  title={Estimation of non-normalized statistical models by score matching.},
  author={Hyv{\"a}rinen, Aapo and Dayan, Peter},
  journal={Journal of Machine Learning Research},
  volume={6},
  number={4},
  year={2005}
}

@book{boyd2011distributed,
  title={Distributed optimization and statistical learning via the alternating direction method of multipliers},
  author={Boyd, Stephen and Parikh, Neal and Chu, Eric},
  year={2011},
  publisher={Now Publishers Inc}
}

@inproceedings{chung2022come,
  title={{Come-Closer-Diffuse-Faster: Accelerating Conditional Diffusion Models for Inverse Problems through Stochastic Contraction}},
  author={Chung, Hyungjin and Sim, Byeongsu and Ye, Jong Chul},
  booktitle={Proceedings of the IEEE/CVF Conference on Computer Vision and Pattern Recognition},
  year={2022}
}

@article{jalal2021robust,
  title={Robust compressed sensing mri with deep generative priors},
  author={Jalal, Ajil and Arvinte, Marius and Daras, Giannis and Price, Eric and Dimakis, Alexandros G and Tamir, Jonathan},
  journal={Advances in Neural Information Processing Systems},
  volume={34},
  year={2021}
}

@article{kim2021noise2score,
  title={{Noise2Score: Tweedie’s Approach to Self-Supervised Image Denoising without Clean Images}},
  author={Kim, Kwanyoung and Ye, Jong Chul},
  journal={Advances in Neural Information Processing Systems},
  volume={34},
  year={2021}
}

@article{zhang2017beyond,
  title={{Beyond a gaussian denoiser: Residual learning of deep CNN for image denoising}},
  author={Zhang, Kai and Zuo, Wangmeng and Chen, Yunjin and Meng, Deyu and Zhang, Lei},
  journal={IEEE transactions on image processing},
  volume={26},
  number={7},
  pages={3142--3155},
  year={2017},
  publisher={IEEE}
}

@article{kawar2021snips,
  title={Snips: Solving noisy inverse problems stochastically},
  author={Kawar, Bahjat and Vaksman, Gregory and Elad, Michael},
  journal={Advances in Neural Information Processing Systems},
  volume={34},
  pages={21757--21769},
  year={2021}
}

@article{anderson1982reverse,
  title={Reverse-time diffusion equation models},
  author={Anderson, Brian DO},
  journal={Stochastic Processes and their Applications},
  volume={12},
  number={3},
  pages={313--326},
  year={1982},
  publisher={Elsevier}
}

@inproceedings{blau2018perception,
  title={The perception-distortion tradeoff},
  author={Blau, Yochai and Michaeli, Tomer},
  booktitle={Proceedings of the IEEE conference on computer vision and pattern recognition},
  pages={6228--6237},
  year={2018}
}

@article{kingma2013auto,
  title={Auto-encoding variational bayes},
  author={Kingma, Diederik P and Welling, Max},
  journal={arXiv preprint arXiv:1312.6114},
  year={2013}
}

@article{efron2011tweedie,
  title={Tweedie’s formula and selection bias},
  author={Efron, Bradley},
  journal={Journal of the American Statistical Association},
  volume={106},
  number={496},
  pages={1602--1614},
  year={2011},
  publisher={Taylor \& Francis}
}

@inproceedings{
song2023pseudoinverseguided,
title={Pseudoinverse-Guided Diffusion Models for Inverse Problems},
author={Jiaming Song and Arash Vahdat and Morteza Mardani and Jan Kautz},
booktitle={International Conference on Learning Representations},
year={2023},
url={https://openreview.net/forum?id=9\_gsMA8MRKQ}
}

@article{
feng2024variational,
title={Variational Bayesian Imaging with an Efficient Surrogate Score-based Prior},
author={Berthy Feng and Katherine Bouman},
journal={Transactions on Machine Learning Research},
issn={2835-8856},
year={2024},
url={https://openreview.net/forum?id=db2pFKVcm1},
note={}
}

@article{mammadov2024amortized,
  title={Amortized posterior sampling with diffusion prior distillation},
  author={Mammadov, Abbas and Chung, Hyungjin and Ye, Jong Chul},
  journal={arXiv preprint arXiv:2407.17907},
  year={2024}
}

@article{chung2025contextmri,
  title={Contextmri: Enhancing compressed sensing MRI through metadata conditioning},
  author={Chung, Hyungjin and Lee, Dohun and Wu, Zihui and Kim, Byung-Hoon and Bouman, Katherine L and Ye, Jong Chul},
  journal={arXiv preprint arXiv:2501.04284},
  year={2025}
}

@article{rozet2024learning,
  title={Learning diffusion priors from observations by expectation maximization},
  author={Rozet, Fran{\c{c}}ois and Andry, G{\'e}r{\^o}me and Lanusse, Fran{\c{c}}ois and Louppe, Gilles},
  journal={Advances in Neural Information Processing Systems},
  volume={37},
  pages={87647--87682},
  year={2024}
}

@article{raphaeli2025silo,
  title={Silo: Solving inverse problems with latent operators},
  author={Raphaeli, Ron and Man, Sean and Elad, Michael},
  journal={arXiv preprint arXiv:2501.11746},
  year={2025}
}

@article{bai2024expectation,
  title={An expectation-maximization algorithm for training clean diffusion models from corrupted observations},
  author={Bai, Weimin and Wang, Yifei and Chen, Wenzheng and Sun, He},
  journal={Advances in Neural Information Processing Systems},
  volume={37},
  pages={19447--19471},
  year={2024}
}

@article{delbracio2023inversion,
  title={Inversion by direct iteration: An alternative to denoising diffusion for image restoration},
  author={Delbracio, Mauricio and Milanfar, Peyman},
  journal={arXiv preprint arXiv:2303.11435},
  year={2023}
}

@article{luo2023image,
  title={Image restoration with mean-reverting stochastic differential equations},
  author={Luo, Ziwei and Gustafsson, Fredrik K and Zhao, Zheng and Sj{\"o}lund, Jens and Sch{\"o}n, Thomas B},
  journal={arXiv preprint arXiv:2301.11699},
  year={2023}
}

@article{liu2023i2sb,
  title={I{$^2$}SB: Image-to-Image Schr{\"o}dinger Bridge},
  author={Liu, Guan-Horng and Vahdat, Arash and Huang, De-An and Theodorou, Evangelos A and Nie, Weili and Anandkumar, Anima},
  journal={arXiv preprint arXiv:2302.05872},
  year={2023},
}

@inproceedings{rout2024beyond,
  title={Beyond first-order tweedie: Solving inverse problems using latent diffusion},
  author={Rout, Litu and Chen, Yujia and Kumar, Abhishek and Caramanis, Constantine and Shakkottai, Sanjay and Chu, Wen-Sheng},
  booktitle={Proceedings of the IEEE/CVF Conference on Computer Vision and Pattern Recognition},
  pages={9472--9481},
  year={2024}
}

@article{chung2023direct,
  title={Direct diffusion bridge using data consistency for inverse problems},
  author={Chung, Hyungjin and Kim, Jeongsol and Ye, Jong Chul},
  journal={Advances in Neural Information Processing Systems},
  volume={36},
  pages={7158--7169},
  year={2023}
}

@article{rout2023solving,
  title={Solving linear inverse problems provably via posterior sampling with latent diffusion models},
  author={Rout, Litu and Raoof, Negin and Daras, Giannis and Caramanis, Constantine and Dimakis, Alex and Shakkottai, Sanjay},
  journal={Advances in Neural Information Processing Systems},
  volume={36},
  pages={49960--49990},
  year={2023}
}

@article{rout2024solving,
  title={Solving linear inverse problems provably via posterior sampling with latent diffusion models},
  author={Rout, Litu and Raoof, Negin and Daras, Giannis and Caramanis, Constantine and Dimakis, Alex and Shakkottai, Sanjay},
  journal={Advances in Neural Information Processing Systems},
  volume={36},
  year={2024}
}

@inproceedings{
    kadkhodaie2021stochastic,
    title={Stochastic Solutions for Linear Inverse Problems using the Prior Implicit in a Denoiser},
    author={Zahra Kadkhodaie and Eero P Simoncelli},
    booktitle={Advances in Neural Information Processing Systems},
    editor={A. Beygelzimer and Y. Dauphin and P. Liang and J. Wortman Vaughan},
    year={2021},
    url={https://openreview.net/forum?id=x5hh6N9bUUb}
}

@article{peng2024improving,
  title={Improving Diffusion Models for Inverse Problems Using Optimal Posterior Covariance},
  author={Peng, Xinyu and Zheng, Ziyang and Dai, Wenrui and Xiao, Nuoqian and Li, Chenglin and Zou, Junni and Xiong, Hongkai},
  journal={arXiv preprint arXiv:2402.02149},
  year={2024}
}

@article{mardani2023variational,
  title={A variational perspective on solving inverse problems with diffusion models},
  author={Mardani, Morteza and Song, Jiaming and Kautz, Jan and Vahdat, Arash},
  journal={arXiv preprint arXiv:2305.04391},
  year={2023}
}

@inproceedings{
    he2024manifold,
    title={Manifold Preserving Guided Diffusion},
    author={Yutong He and Naoki Murata and Chieh-Hsin Lai and Yuhta Takida and Toshimitsu Uesaka and Dongjun Kim and Wei-Hsiang Liao and Yuki Mitsufuji and J Zico Kolter and Ruslan Salakhutdinov and Stefano Ermon},
    booktitle={The Twelfth International Conference on Learning Representations},
    year={2024},
    url={https://openreview.net/forum?id=o3BxOLoxm1}
}

@article{yang2024guidance,
  title={Guidance with Spherical Gaussian Constraint for Conditional Diffusion},
  author={Yang, Lingxiao and Ding, Shutong and Cai, Yifan and Yu, Jingyi and Wang, Jingya and Shi, Ye},
  journal={arXiv preprint arXiv:2402.03201},
  year={2024}
}

@inproceedings{
    liu2023flow,
    title={Flow Straight and Fast: Learning to Generate and Transfer Data with Rectified Flow},
    author={Xingchao Liu and Chengyue Gong and qiang liu},
    booktitle={The Eleventh International Conference on Learning Representations },
    year={2023},
    url={https://openreview.net/forum?id=XVjTT1nw5z}
}

@inproceedings{
    lipman2023flow,
    title={Flow Matching for Generative Modeling},
    author={Yaron Lipman and Ricky T. Q. Chen and Heli Ben-Hamu and Maximilian Nickel and Matthew Le},
    booktitle={The Eleventh International Conference on Learning Representations },
    year={2023},
    url={https://openreview.net/forum?id=PqvMRDCJT9t}
}

@article{daras2025ambient,
  title={Ambient Diffusion Omni: Training Good Models with Bad Data},
  author={Daras, Giannis and Rodriguez-Munoz, Adrian and Klivans, Adam and Torralba, Antonio and Daskalakis, Constantinos},
  journal={arXiv preprint arXiv:2506.10038},
  year={2025}
}

@inproceedings{
chung2024prompttuning,
title={Prompt-tuning Latent Diffusion Models for Inverse Problems},
author={Hyungjin Chung and Jong Chul Ye and Peyman Milanfar and Mauricio Delbracio},
booktitle={Forty-first International Conference on Machine Learning},
year={2024},
url={https://openreview.net/forum?id=hrwIndai8e}
}

@inproceedings{
kim2025regularization,
title={Regularization by Texts for Latent Diffusion Inverse Solvers},
author={Jeongsol Kim and Geon Yeong Park and Hyungjin Chung and Jong Chul Ye},
booktitle={The Thirteenth International Conference on Learning Representations},
year={2025},
url={https://openreview.net/forum?id=TtUh0TOlGX}
}

@article{daras2024survey,
  title={A survey on diffusion models for inverse problems},
  author={Daras, Giannis and Chung, Hyungjin and Lai, Chieh-Hsin and Mitsufuji, Yuki and Ye, Jong Chul and Milanfar, Peyman and Dimakis, Alexandros G and Delbracio, Mauricio},
  journal={arXiv preprint arXiv:2410.00083},
  year={2024}
}

@article{xu2025rethinking,
  title={Rethinking Diffusion Posterior Sampling: From Conditional Score Estimator to Maximizing a Posterior},
  author={Xu, Tongda and Cai, Xiyan and Zhang, Xinjie and Ge, Xingtong and He, Dailan and Sun, Ming and Liu, Jingjing and Zhang, Ya-Qin and Li, Jian and Wang, Yan},
  journal={arXiv preprint arXiv:2501.18913},
  year={2025}
}

@inproceedings{wu2024diffusion,
  title={Diffusion Posterior Proximal Sampling for Image Restoration},
  author={Wu, Hongjie and He, Linchao and Zhang, Mingqin and Chen, Dongdong and Luo, Kunming and Luo, Mengting and Zhou, Ji-Zhe and Chen, Hu and Lv, Jiancheng},
  booktitle={Proceedings of the 32nd ACM International Conference on Multimedia},
  pages={214--223},
  year={2024}
}

@inproceedings{zhang2025improving,
  title={Improving diffusion inverse problem solving with decoupled noise annealing},
  author={Zhang, Bingliang and Chu, Wenda and Berner, Julius and Meng, Chenlin and Anandkumar, Anima and Song, Yang},
  booktitle={Proceedings of the Computer Vision and Pattern Recognition Conference},
  pages={20895--20905},
  year={2025}
}

@article{li2024decoupled,
  title={Decoupled data consistency with diffusion purification for image restoration},
  author={Li, Xiang and Kwon, Soo Min and Liang, Shijun and Alkhouri, Ismail R and Ravishankar, Saiprasad and Qu, Qing},
  journal={arXiv preprint arXiv:2403.06054},
  year={2024}
}

@article{alkhouri2024sitcom,
  title={SITCOM: Step-wise Triple-Consistent Diffusion Sampling for Inverse Problems},
  author={Alkhouri, Ismail and Liang, Shijun and Huang, Cheng-Han and Dai, Jimmy and Qu, Qing and Ravishankar, Saiprasad and Wang, Rongrong},
  journal={arXiv preprint arXiv:2410.04479},
  year={2024}
}

@book{tarantola2005inverse,
  title={Inverse problem theory and methods for model parameter estimation},
  author={Tarantola, Albert},
  year={2005},
  publisher={SIAM}
}

@article{hu2024learning,
  title={Learning image priors through patch-based diffusion models for solving inverse problems},
  author={Hu, Jason and Song, Bowen and Xu, Xiaojian and Shen, Liyue and Fessler, Jeffrey A},
  journal={Advances in Neural Information Processing Systems},
  volume={37},
  pages={1625--1660},
  year={2024}
}

@article{stein1981estimation,
  title={Estimation of the mean of a multivariate normal distribution},
  author={Stein, Charles M},
  journal={The annals of Statistics},
  pages={1135--1151},
  year={1981},
  publisher={JSTOR}
}

@article{eldar2008generalized,
  title={Generalized SURE for exponential families: Applications to regularization},
  author={Eldar, Yonina C},
  journal={IEEE Transactions on Signal Processing},
  volume={57},
  number={2},
  pages={471--481},
  year={2008},
  publisher={IEEE}
}

@inproceedings{
daras2024consistent,
title={Consistent Diffusion Meets Tweedie: Training Exact Ambient Diffusion Models with Noisy Data},
author={Giannis Daras and Alex Dimakis and Constantinos Costis Daskalakis},
booktitle={Forty-first International Conference on Machine Learning},
year={2024},
url={https://openreview.net/forum?id=PlVjIGaFdH}
}

@article{daras2023consistent,
  title={Consistent diffusion models: Mitigating sampling drift by learning to be consistent},
  author={Daras, Giannis and Dagan, Yuval and Dimakis, Alex and Daskalakis, Constantinos},
  journal={Advances in Neural Information Processing Systems},
  volume={36},
  pages={42038--42063},
  year={2023}
}

@inproceedings{song2023consistency,
  title={Consistency models},
  author={Song, Yang and Dhariwal, Prafulla and Chen, Mark and Sutskever, Ilya},
  booktitle={Proceedings of the 40th International Conference on Machine Learning},
  pages={32211--32252},
  year={2023}
}

@article{
kawar2024gsurebased,
title={{GSURE}-Based Diffusion Model Training with Corrupted Data},
author={Bahjat Kawar and Noam Elata and Tomer Michaeli and Michael Elad},
journal={Transactions on Machine Learning Research},
issn={2835-8856},
year={2024},
url={https://openreview.net/forum?id=BRl7fqMwaJ},
note={}
}

@article{aggarwal2022ensure,
  title={ENSURE: A general approach for unsupervised training of deep image reconstruction algorithms},
  author={Aggarwal, Hemant Kumar and Pramanik, Aniket and John, Maneesh and Jacob, Mathews},
  journal={IEEE transactions on medical imaging},
  volume={42},
  number={4},
  pages={1133--1144},
  year={2022},
  publisher={IEEE}
}

@article{hu2025test,
  title={Test-Time Adaptation Improves Inverse Problem Solving with Patch-Based Diffusion Models},
  author={Hu, Jason and Song, Bowen and Fessler, Jeffrey A and Shen, Liyue},
  journal={IEEE Transactions on Computational Imaging},
  year={2025},
  publisher={IEEE}
}

@article{daras2023ambient,
  title={Ambient diffusion: Learning clean distributions from corrupted data},
  author={Daras, Giannis and Shah, Kulin and Dagan, Yuval and Gollakota, Aravind and Dimakis, Alex and Klivans, Adam},
  journal={Advances in Neural Information Processing Systems},
  volume={36},
  pages={288--313},
  year={2023}
}

@inproceedings{gao2025diffusionmeetsflow,
  author = {Gao, Ruiqi and Hoogeboom, Emiel and Heek, Jonathan and Bortoli, Valentin De and Murphy, Kevin P. and Salimans, Tim},
  title = {Diffusion Meets Flow Matching: Two Sides of the Same Coin},
  year = {2024},
  url  = {https://diffusionflow.github.io/}
}

@article{akiyama2019first,
  title={First M87 event horizon telescope results. IV. Imaging the central supermassive black hole},
  author={Akiyama, Kazunori and Alberdi, Antxon and Alef, Walter and Asada, Keiichi and Azulay, Rebecca and Baczko, Anne-Kathrin and Ball, David and Balokovi{\'c}, Mislav and Barrett, John and Bintley, Dan and others},
  journal={The Astrophysical Journal Letters},
  volume={875},
  number={1},
  pages={L4},
  year={2019},
  publisher={IoP Publishing}
}

@article{gupta2021cryogan,
  title={CryoGAN: A new reconstruction paradigm for single-particle cryo-EM via deep adversarial learning},
  author={Gupta, Harshit and McCann, Michael T and Donati, Laurene and Unser, Michael},
  journal={IEEE Transactions on Computational Imaging},
  volume={7},
  pages={759--774},
  year={2021},
  publisher={IEEE}
}

@article{barbano2025steerable,
  title={Steerable Conditional Diffusion for Out-of-Distribution Adaptation in Medical Image Reconstruction},
  author={Barbano, Riccardo and Denker, Alexander and Chung, Hyungjin and Roh, Tae Hoon and Arridge, Simon and Maass, Peter and Jin, Bangti and Ye, Jong Chul},
  journal={IEEE Transactions on Medical Imaging},
  year={2025},
  publisher={IEEE}
}

@inproceedings{chung2024deep,
  title={Deep diffusion image prior for efficient ood adaptation in 3d inverse problems},
  author={Chung, Hyungjin and Ye, Jong Chul},
  booktitle={European Conference on Computer Vision},
  pages={432--455},
  year={2024},
  organization={Springer}
}

@inproceedings{ulyanov2018deep,
  title={Deep image prior},
  author={Ulyanov, Dmitry and Vedaldi, Andrea and Lempitsky, Victor},
  booktitle={Proceedings of the IEEE conference on computer vision and pattern recognition},
  pages={9446--9454},
  year={2018}
}

@inproceedings{
zilberstein2025repulsive,
title={Repulsive Latent Score Distillation for Solving Inverse Problems},
author={Nicolas Zilberstein and Morteza Mardani and Santiago Segarra},
booktitle={The Thirteenth International Conference on Learning Representations},
year={2025},
url={https://openreview.net/forum?id=bwJxUB0y46}
}

@inproceedings{feng2023score,
  title={Score-based diffusion models as principled priors for inverse imaging},
  author={Feng, Berthy T and Smith, Jamie and Rubinstein, Michael and Chang, Huiwen and Bouman, Katherine L and Freeman, William T},
  booktitle={Proceedings of the IEEE/CVF International Conference on Computer Vision},
  pages={10520--10531},
  year={2023}
}

@inproceedings{murata2023gibbsddrm,
  title={Gibbsddrm: A partially collapsed gibbs sampler for solving blind inverse problems with denoising diffusion restoration},
  author={Murata, Naoki and Saito, Koichi and Lai, Chieh-Hsin and Takida, Yuhta and Uesaka, Toshimitsu and Mitsufuji, Yuki and Ermon, Stefano},
  booktitle={International conference on machine learning},
  pages={25501--25522},
  year={2023},
  organization={PMLR}
}

@article{man2025proxies,
  title={Proxies for Distortion and Consistency with Applications for Real-World Image Restoration},
  author={Man, Sean and Ohayon, Guy and Raphaeli, Ron and Elad, Michael},
  journal={arXiv preprint arXiv:2501.12102},
  year={2025}
}

@inproceedings{laroche2024fast,
  title={Fast diffusion em: a diffusion model for blind inverse problems with application to deconvolution},
  author={Laroche, Charles and Almansa, Andr{\'e}s and Coupete, Eva},
  booktitle={Proceedings of the IEEE/CVF Winter Conference on Applications of Computer Vision},
  pages={5271--5281},
  year={2024}
}

@inproceedings{venkatakrishnan2013plug,
  title     = {Plug-and-Play Priors for Model Based Reconstruction},
  author    = {Singanallur V. Venkatakrishnan and Charles A. Bouman and Rebecca M. Willett},
  booktitle = {2013 IEEE Global Conference on Signal and Information Processing (GlobalSIP)},
  pages     = {945--948},
  year      = {2013},
  organization = {IEEE}
}

@inproceedings{lee2024diffusion,
  title={Diffusion prior-based amortized variational inference for noisy inverse problems},
  author={Lee, Sojin and Park, Dogyun and Kong, Inho and Kim, Hyunwoo J},
  booktitle={European Conference on Computer Vision},
  pages={288--304},
  year={2024},
  organization={Springer}
}

@article{romano2017little,
  title={The little engine that could: Regularization by denoising (RED)},
  author={Romano, Yaniv and Elad, Michael and Milanfar, Peyman},
  journal={SIAM journal on imaging sciences},
  volume={10},
  number={4},
  pages={1804--1844},
  year={2017},
  publisher={SIAM}
}

@inproceedings{
dinh2017density,
title={Density estimation using Real {NVP}},
author={Laurent Dinh and Jascha Sohl-Dickstein and Samy Bengio},
booktitle={International Conference on Learning Representations},
year={2017},
url={https://openreview.net/forum?id=HkpbnH9lx}
}

@article{erbach2025solving,
  title={Solving Inverse Problems with FLAIR},
  author={Erbach, Julius and Narnhofer, Dominik and Dombos, Andreas and Schiele, Bernt and Lenssen, Jan Eric and Schindler, Konrad},
  journal={arXiv preprint arXiv:2506.02680},
  year={2025}
}

@article{kim2025flowdps,
  title={Flowdps: Flow-driven posterior sampling for inverse problems},
  author={Kim, Jeongsol and Kim, Bryan Sangwoo and Ye, Jong Chul},
  journal={arXiv preprint arXiv:2503.08136},
  year={2025}
}

@article{vahdat2021score,
  title={Score-based generative modeling in latent space},
  author={Vahdat, Arash and Kreis, Karsten and Kautz, Jan},
  journal={Advances in neural information processing systems},
  volume={34},
  pages={11287--11302},
  year={2021}
}

@article{patel2024steering,
  title={Steering rectified flow models in the vector field for controlled image generation},
  author={Patel, Maitreya and Wen, Song and Metaxas, Dimitris N and Yang, Yezhou},
  journal={arXiv preprint arXiv:2412.00100},
  year={2024}
}
